\documentclass{article}
\usepackage{Examplee}
\usepackage{graphicx}
\usepackage{setspace}
\usepackage{graphicx}
\usepackage{amssymb}
\usepackage{amsmath} 
\usepackage{epstopdf}
\usepackage{subfig}

\usepackage{float}
\usepackage{tabulary}
\usepackage{tabularx}
\usepackage{verbatim} 
\usepackage{geometry}                
\usepackage{hyperref}
\hypersetup{colorlinks,breaklinks,
linkcolor=[rgb]{0.5,0.,0.},
citecolor=[rgb]{0.000,0.427,0.173},
urlcolor=[rgb]{0.031,0.318,0.612}}
\usepackage{fancyhdr}

\usepackage{todonotes}
\usepackage{times} 
\usepackage{multicol}
\usepackage{tensor}
\usepackage{multirow}
\usepackage{booktabs}

\def\fps@figure{htp}
\def\fps@table{htp}

\newcommand{\bi}{\begin{itemize}}
\newcommand{\ei}{\end{itemize}}

\newcommand{\bfig}{\begin{figure}}
\newcommand{\efig}{\end{figure}}

\newcommand{\benum}{\begin{enumerate}}
\newcommand{\eenum}{\end{enumerate}}

\newcommand{\be}{\begin{equation}}
\newcommand{\ee}{\end{equation}}

\newcommand{\ba}{\begin{eqnarray}}
\newcommand{\ea}{\end{eqnarray}}

\newcommand{\etal}{{et al.}}

\renewcommand{\vec}[1]{\underline{#1}}

\newcommand{\unit}[1]{\mbox{$\rm \,#1$}}

\usepackage{color}

\definecolor{CommentRed}{rgb}{0.7,0,0}
\definecolor{CommentBlue}{rgb}{0,0,0.7}
\definecolor{CommentDG}{rgb}{0,0.6,0}

\newcommand{\coordi}[1]{$\{\mathcal{#1}\}$}

\newenvironment{Contfigure}{%
\begin{figure}}{%
\end{figure}}

\title{Build your own visual-inertial odometry aided cost-effective and open-source autonomous drone}
\author{
Inkyu $\text{Sa}^{*}$, Mina Kamel, Michael Burri, Michael Bloesch,  \AND Raghav Khanna, Marija Popovi\'{c}, Juan Nieto, and Roland Siegwart \\ \\Autonomous Systems Lab., ETH Zurich\\
Z\"{u}rich, 8092, Switzerland\\${}^*$\texttt{inkyu.sa@mavt.ethz.ch}}
\begin{document}

\maketitle

\begin{abstract}
This paper describes an approach to building a cost-effective and research grade visual-inertial odometry aided vertical taking-off and landing (VTOL) platform. We utilize an off-the-shelf visual-inertial sensor, an onboard computer, and a quadrotor platform that are factory-calibrated and mass-produced, thereby sharing similar hardware and sensor specifications (e.g., mass, dimensions, intrinsic and extrinsic of camera-IMU systems, and signal-to-noise ratio). We then perform system calibration and identification enabling the use of our visual-inertial odometry, multi-sensor fusion, and model predictive control frameworks with the off-the-shelf products. This approach partially circumvents the tedious parameter tuning procedures required to build a full system. The complete system is evaluated extensively both indoors using a motion capture system and outdoors using a laser tracker while performing hover and step responses, and trajectory following tasks in the presence of external wind disturbances. We achieve root-mean-square (RMS) pose errors of 0.036\unit{m} with respect to reference hover trajectories. We also conduct relatively long distance ($\approx$180\unit{m}) experiments on a farm site, demonstrating 0.82\% drift error of the total flight distance. This paper conveys the insights we acquired about the platform and sensor module and offers open-source code with tutorial documentation to the community. 
\end{abstract}

\section{INTRODUCTION}
\label{sec:intro}
Over the past decade, vertical taking-off and landing (VTOL) micro aerial vehicles (MAVs), which use counter-rotating rotors to generate thrusting and rotational forces, have gained renown in both research and industry.
Emerging applications, including structural inspection, aerial photography, cinematography, and environmental surveillance, have spurred a wide range of ready-to-fly commercial platforms,
whose performance has improved steadily in terms of flight time, payload, and safety-related smart-features, allowing for more stable, easier and safer pilot maneuvers.
However, a key challenge is directly adapting commercial platforms \cite{lim2012build} for tasks requiring accurate dynamic models, high-performance controllers, and precise, low-latency state estimators, such as obstacle avoidance and path planning \cite{burri2015real,Nuske:2015aa,Yang-:2015aa,oleynikova2016continuous-time}, landing on moving platforms \cite{Lee:2012aa}, object picking \cite{mellinger2011design}, and precision agriculture \cite{Zhang:2012aa}.

Research-grade MAV platforms can alleviate these issues. For instance, Ascending Technologies provides excellent products \cite{Achtelik:2011fk,Weiss:2011aa} for advanced aerial applications \cite{hwangbo2017control} with an accompanying software development kit (SDK), scientific resources (including self-contained documentation), and online support. 
However, they are relatively expensive for research purposes, and part replacements are difficult due to limited retail services.

\begin{figure}
\begin{center}
\includegraphics[width=0.7\columnwidth]{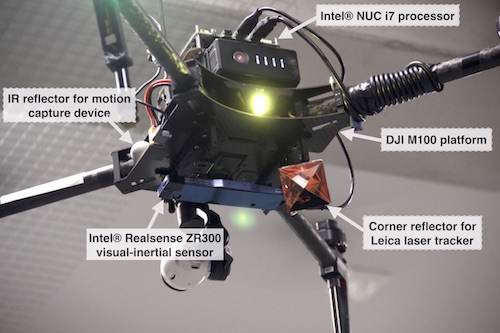}
\end{center}
    \caption{Matrice 100 VTOL MAV quadrotor platform with a downward-facing VI sensor. The IR reflectors and prism are mounted for obtaining ground truth.}
    \label{fig:M100}
\end{figure}

For VTOL MAVs to become more widespread, they require lower costs and more easily obtainable parts. Recently, the DJI Matrice 100 MAV\footnote{https://www.dji.com/matrice100} (Fig.~\ref{fig:M100}) has been introduced as a commercial platform with parts available at local retailers. Developers can access sensor data, e.g., from the inertial measurement unit (IMU) and barometers, and send commands to the low-level attitude controller through the SDK \cite{borowczyk2016autonomous}. Although manufacturer documentation is provided, there are limited scientific resources detailing aspects such as attitude dynamics, the underlying autopilot controller structure, and input command scaling. This information is critical for subsequent position control strategies such as model predictive control (MPC).

A visual-inertial (VI) sensor is a favorable choice for aerial robotics due to its light weight, low power consumption, and ability to recover unknown scales (monocular camera). This sensor suite can provide time-synchronized wide field of view (FoV) image ($\approx133^{\circ}$) and motion measurements. There is an impressive research-grade visual-inertial sensor \cite{nikolic2014synchronized} providing time-synchronized and calibrated IMU-stereo camera images with supporting documentation and device drivers. Unfortunately, this sensor is relatively expensive and its production was discontinued. The Intel ZR300\footnote{http://click.intel.com/realsense.html} has recently emerged as a compelling alternative. It is an affordable and mass-produced module that enables applying the same configuration and calibration parameters to other sensors with low performance penalties. Table~\ref{tbl:sum} summarizes the total cost for the VI sensor, MAV, and PC.

\begin{table}[]
\centering
\caption{Summary of total system cost}
\label{tbl:sum}
\begin{tabular}{ccc}
\hline
\textbf{ID}  & \textbf{Name}                        & \textbf{Price (USD)} \\ \hline
1   & M100\footnotemark                        & 1979.4  \\ \hline
2   & Visual-inertial sensor      & 289     \\ \hline
3   & Intel NUC i7                & 482     \\ \hline
4   & 1TB SSD                     & 280     \\ \hline
5   & 16 GB RAM                   & 103.5   \\ \hline
6   & TTL to USB converter (FTDI) & 14.75   \\ \hline
7   & Power regulator (12V)       & 21      \\ \hline
\textbf{Sum} &                             & \textbf{3169.65}
\end{tabular}
\end{table}
\footnotetext{You can get 40\% discount with developer registration that we have to create to activate DJI's Onboard SDK (Please have a look the above link for the registration). Otherwise, we can't send commands to the N1 autopilot. We thus list 40\% discounted price in the part list.}

In this paper, we address these gaps by using a ready-to-market affordable VTOL platform and a VI sensor to perform system identification, sensor calibration, and system integration. The system can autonomously track motion commands with high precision in indoor and outdoor environments. This ability is fundamental for any high-level application, enabling researchers to build custom systems without tedious tuning procedures. Moreover, we provide self-contained documentation to cater for set-ups with different inertial moments and sensor mount configurations. The contributions of this system paper are:

\begin{itemize}
\item a delivery of software packages (including a modified SDK, nonlinear MPC, calibration parameters, and system identification tools) with accompanying documentation to the community,
\item an evaluation of control and state estimation performance in different environments,
\item a demonstration of use-cases adapting our approach to build custom research platforms with a step-by-step tutorial available at:
\end{itemize}

\begin{center}\texttt{https://goo.gl/yj8WsZ}\end{center}

The techniques presented are independent from the platforms used in this paper. For example, the MPC strategy is applicable to other commercial VTOL platforms (e.g., Matrice 200 or 600) given identified dynamic system models using our procedures. The Robust Visual-Inertial Odometry (ROVIO) framework can also be utilized in mobile platforms for navigation tasks \cite{siegwart2011introduction} with the VI sensor (Intel ZR300).

The remainder of this paper is structured as follows. Section \ref{sec:background} introduces the state-of-the-art in MAV VI odometry, dynamic system identification, and control. Sections \ref{sec:methodologies} and \ref{sec:rovio} detail the vehicle specification, and VI odometry framework used in this work. Our system identification and control strategies are described in Section \ref{sec:dynamics}. We present experimental results in Section \ref{sec:results} before concluding in section \ref{sec:conclusion}.
\section{Related Work}\label{sec:background}
VTOL MAVs are used increasingly for tasks such as building inspection, aerial photography, and precision agriculture. To perform these missions, capabilities of localization, perception, control, and path planning are critical. These topics are very broad and it is challenging to cover them comprehensively in this article. We thus focus on the state-of-the-art VI odometry techniques, dynamic system identification, and VTOL MAV control as most relevant sub-topics.

VI odometry has been an active research topic in the last decade given advances in microelectromechanical systems (MEMS), IMU, and imaging technologies \cite{scaramuzza2011visual}. It is a favorable option for payload-constrained platforms such as VTOL MAVs due to its light weight and relatively low computational requirements. All sensor states are jointly-coupled (i.e., tightly-coupled) within (i) filtering or (ii) nonlinear optimization frameworks.
These frameworks estimate a robot's ego-motion by integrating visual and inertial measurements while minimizing data preprocessing and thereby improving stochastic consistency.
Filter-based approaches often do this within a Kalman Filter (e.g. \cite{Bloesch:2015aa}), thus suffering from drift and exhibiting only a limited representation of the global environment.
However, the resulting localization accuracy suffices for stabilizing control and executing local tasks.
Furthermore, these frameworks procure estimates which can be input to feedback control strategies at a constant computational cost. In contrast, the second class of methods performs keyframe-based nonlinear optimization over all jointly-coupled sensor states \cite{leutenegger2015keyframe,forster2015manifold,lynen2015get,shen2014initialization,mur2017visual}. In some cases, these approaches can also perform loop closures to compensate for drift and provide globally consistent maps. However, they often demand costly computations that may burden smaller platforms, and demonstrate state estimation without feedback control (or only for hovering tasks).

Identifying the dynamics 
of attitude controllers is vital to achieving good control performance.
For a common quadrotor, the rigid vehicle dynamics are well-known \cite{bouabdallah2007design,corke2011robotics} and can be modeled as a non-linear system with individual rotors attached to a rigid airframe, accounting for drag force and blade flapping~\cite{mahony2012multirotor}. However, the identification of attitude controllers is often non-trivial for consumer products due to limited scientific resources and thus requires techniques to estimate dynamic model parameters. Traditionally, parameter estimation is performed offline using complete measurement data obtained from a physical test bed and CAD models~\cite{Pounds:2009fk,hoffmann2008quadrotor}. A linear least-squares method is used to estimate parameters from recorded data in batch offline processing \cite{Sa:2012ICRA,tischler2006aircraft}. This approach only requires flight datasets; however, identification must be repeated if the vehicle configuration changes. In contrast, online system identification involves applying recursive estimation to real-time sensor data. Burri \etal\;\cite{burri2016maximum} present a method for identifying the dominant dynamic parameters of a VTOL MAV using the maximum likelihood approach, and apply it to estimate moments of inertia and aerodynamic parameters.
We follow a batch-based strategy to determine the dynamic vehicle parameters from short manual piloting maneuvers. This allows us to obtain the parameters needed for MPC \cite{kamelmpc2016} using only the onboard IMU and without restrictive simplifying assumptions.



\section{VTOL MAV platform and visual-inertial sensor module}\label{sec:methodologies}
This section describes the vehicle (Matrice 100) and sensor module (Intel ZR300) used in this paper. Their general specifications are well-documented and available from official sources. Here, we only highlight the information relevant for building research platforms, including auto-trim compensation, dead-zone recovery, camera-IMU extrinsic calibration, and their time synchronization.

\subsection{VTOL MAV platform}\label{sec:M100}
Our vehicle is a quadrotor with 650\unit{mm} diagonal length, and four 13\unit{in} diameter propellers with 4.5\unit{in} thread pitch. The maximum takeoff weight is 3600\unit{g} and flight time varies depending on the hardware configuration (16$\sim$40\unit{min}). The N1 autopilot manages attitude control and telemetry, but information regarding the device is not publicly disclosed. Using the SDK, sensor data can be accessed through serial communication, and we configure the IMU update rate at 50\unit{Hz}. The SDK enables access to most functionalities and supports cross-platform development environments such as the Robot Operating System (ROS), Android, and iOS. However, there is a fundamental issue in sending control commands with this protocol. The manufacturer uses \texttt{ROS services} to send commands; this is strongly not recommended\footnote{http://wiki.ros.org/ROS/Patterns/Communication} as they are blocking calls that should be used only for triggering signals or quick calculations. If data transaction (hand-shaking) fails (e.g., due to poor WiFi signal), it blocks all subsequent calls. Given that low control command latency ($\approx$ 20\unit{ms}) can have large performance impacts, we modify the SDK to send direct control commands through the serial port. It is worth mentioning that we faced communication problems (921600\unit{bps}) while receiving/sending data at 100\unit{Hz}, as the onboard computer could not transmit any commands to the N1 autopilot at this particular frequency. Consequently, we used 50\unit{Hz} in all presented experiments, and are currently investigating this issue.

There are usually trim switches on an ordinary transmitter that allow for small command input adjustments. The N1 autopilot, however, has an auto-trim feature that balances attitude by estimating horizontal velocity. This permits easier and safer manual piloting but introduces a constant position error offset for autonomous control. To address this, we estimate the balancing point where the vehicle's motion is minimum (hovering) and adjust the neutral position to this point. If there is a change in an inertial moment (e.g., mounting a new device or changing the battery position), the balancing position must be updated.

Another interesting autopilot feature is the presence of a dead zone in the small range close to the neutral value where it ignores all input commands. This function is also useful for manual piloting since the vehicle should ignore perturbations from hand tremors, but it significantly degrades control performance. We determine this range by sweeping control commands around the neutral and detecting the control inputs when any motion is detected (i.e., horizontal and vertical velocity changes). As this task is difficult with a real VTOL platform due to its fast and naturally unstable dynamics, we use the hardware-in-loop simulator enabling input command reception from the transmitter. If the commands are within those ranges, they are set as the maximum/minimum dead zone values.

\subsection{Visual-inertial sensor}\label{sec:rz300}
\begin{figure}
\center
\includegraphics[width=0.9\columnwidth]{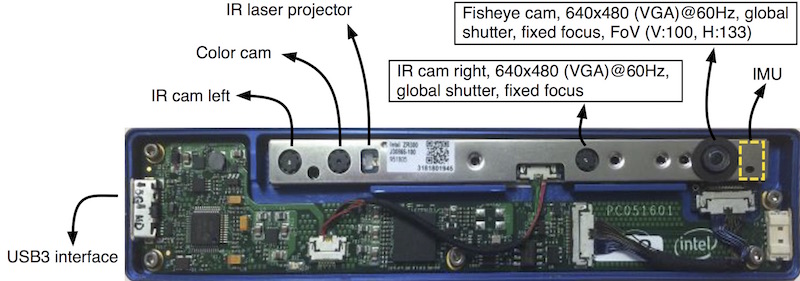}
\caption{Intel ZR300 VI sensor module.}
\label{fig:zr300}
\end{figure}

In this paper, we exploit a ready-to-market VI sensor\footnote{http://click.intel.com/intelr-realsensetm-development-kit-featuring-the-zr300.html}. Most importantly, the sensor has one fisheye camera with a FoV of $133^{\circ}$ and $100^{\circ}$ in the horizontal and vertical directions, respectively, and streams a 640$\times$480 image at 60 frames per second shown in Fig. \ref{fig:zr300}. An onboard IMU provides 3-axis accelerations and angular velocities in a body frame with time stamps at 20\unit{kHz}. As we do not use depth measurements, the two IR cameras, the projector, and the RGB camera, they are disabled for reduced-power operation.

Camera-IMU time synchronization for any VI related tasks is non-trivial since camera images and motions measured by the IMU are tightly connected. In the following section, we introduce the time synchronization and camera-IMU extrinsic calibration.

\subsubsection{Camera-IMU time-synchronization}\label{sec:time_sync}
\begin{figure}
\centering
\includegraphics[width=0.8\columnwidth]{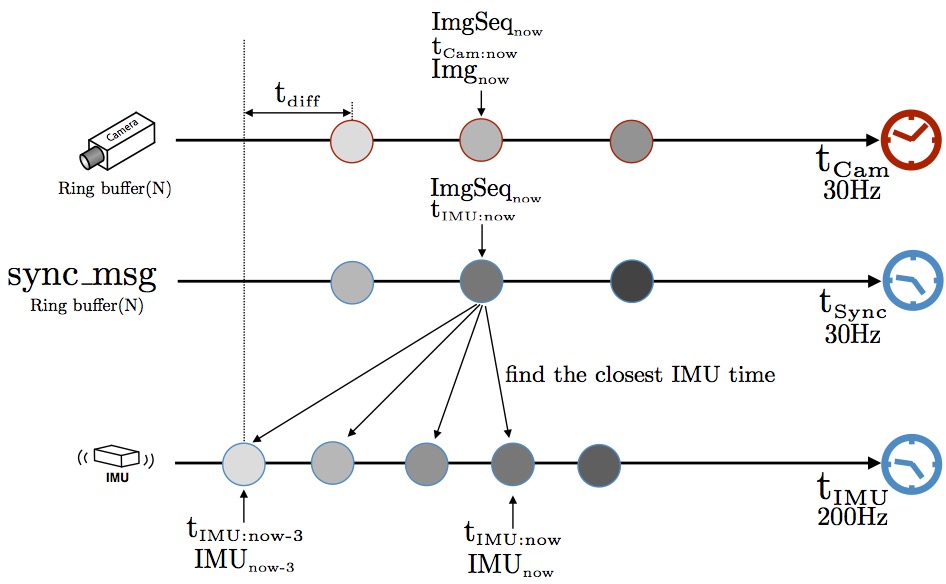}
\caption{Camera-IMU time-synchronization illustration. Two sensors are running at different rates with own time sources as indicated by the red and blue clocks. The faster update rate of the IMU, $\text{t}_{\mbox{\tiny IMU:now}}$ is used as the master time. For each node, the shade of gray represents amount of time elapsed (i.e., the same shade signifies the same time). $\text{N}$ denotes the size of the ring buffers.}
\label{fig:timeSync}
\end{figure}

The ZR300 has different clock sources for the IMU and image timestamps.
Therefore, direct usage of the timestamps from the RealSense library leads to poor estimator performance.
To mitigate this issue, a synchronization message is generated every time the sensor captures an image,
which contains the timestamp and sequence number of the corresponding image.
The same sequence number is also contained in the image message.
We implemented two ring buffers for images and synchronization messages to look up the correct timestamp of the image with respect to the IMU before publishing it over ROS. This procedure is depicted in Fig.~\ref{fig:timeSync}.

Another important aspect is that the sensor IMU has different sampling rates on its gyroscopes ($\sim$200\unit{Hz}) and accelerometers ($\sim$250\unit{Hz}).
Since the noise density of the gyroscopes is smaller than those of the accelerometers and the data is more important for state estimation, we publish an IMU message containing both sensors at the rate of the gyroscopes.
This requires buffering the messages and linear interpolation of the accelerometer messages.

In our present version of the sensor, the IMU is not intrinsically calibrated.
We use the extended version of Kalibr \cite{rehder2016extending} to estimate each axis of the gyro and the accelerometer with respect to a reference. 

Lastly, we currently do not compensate for the camera exposure time.
The timestamps are triggered at the beginning of the exposure time rather than in the middle.
This could be important in cases of large lighting changes.
Instead, we use a constant offset between the IMU and camera as estimated using Kalibr \cite{furgale2013unified}.

\subsection{Coordinate systems definition}
\begin{figure}
\centering
\includegraphics[width=0.8\columnwidth]{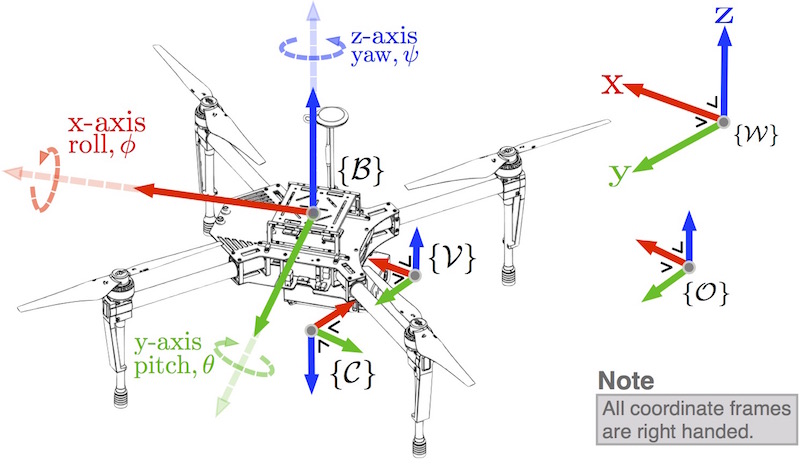}
\caption{Coordinate system definitions\protect\footnotemark. There are 5 frames: world $\mathcal{W}$, odometry $\mathcal{O}$, body $\mathcal{B}$, camera $\mathcal{C}$, and VI sensor IMU $\mathcal{V}$.}
\label{fig:coordi}
\end{figure}

We define 5 right-handed frames following standard ROS convention: world $\{\mathcal{W}\}$, odometry $\{\mathcal{O}\}$, body $\{\mathcal{B}\}$, camera $\{\mathcal{C}\}$, and VI sensor IMU $\{\mathcal{V}\}$ as shown in Fig \ref{fig:coordi}. $\{\mathcal{B}\}$ is aligned with the vehicle's IMU frame and its x-axis indicates the forward direction of the vehicle, with the y- and z-axes as left and up, respectively. We use Euler angles; roll $\phi$, pitch $\theta$, and yaw $\psi$ about the x, y, and z-axis respectively for Root Mean Square (RMS) error calculation and visualization. Quaternions are utilized for any computational processes. Note that the default vehicle coordinate system configuration is North-East-Down, so that the angle, acceleration, and angular velocity measurements from the onboard IMU are rotated $\pi$ along the x-axis to align with $\mathcal{B}$. The $\{\mathcal{W}\}$ and $\{\mathcal{O}\}$ frames are fixed coordinate where VI odometry is initialized. We treat them identically in this paper, but they can differ if external pose measurements (e.g., GPS or Leica laser tracker) are employed. 

These coordinate systems and notations are used in the rest of this paper.
\footnotetext{Image source from http://goo.gl/7NsbmG}

\subsubsection{Extrinsic calibration}\label{sec:kalibr}
\begin{figure}
\centering
\includegraphics[width=0.5\columnwidth]{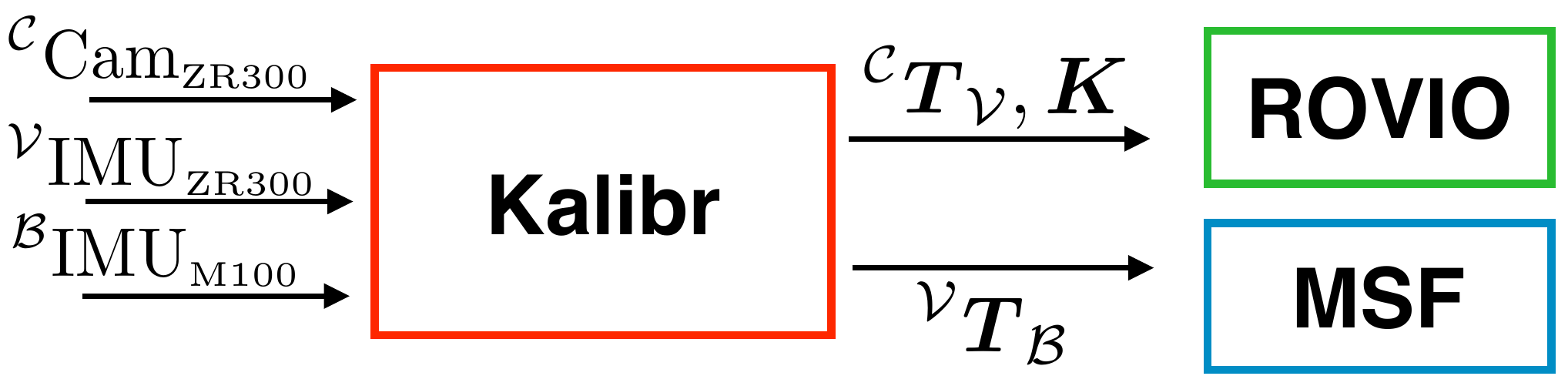}
\caption{Three inputs of image and IMUs are fed into the Kalibr calibration framework. Intrinsic ($\boldsymbol{K}$) and extrinsic (${}^{\mathcal{C}}\boldsymbol{T}_{\mathcal{V}}$ and ${}^{\mathcal{V}}\boldsymbol{T}_{\mathcal{B}}$) are utilized by subsequent VI odometry and sensor fusion frameworks.}
\label{fig:kalibr}
\end{figure}

Two essential transformations are required before a flight; 1) the transformation from the vehicle IMU frame \coordi{B} to the VI sensor's camera frame \coordi{C}, i.e., ${}^{\mathcal{B}}\boldsymbol{T}_{\mathcal{C}}\in SE(3)\subset\mathbb{R}^{\mbox{\tiny 4$\times$4}}$, and 2) the transformation from the camera frame \coordi{C} to VI sensor's IMU frame \coordi{V}, ${}^{\mathcal{C}}\boldsymbol{T}_{\mathcal{V}}$. These extrinsic parameters and camera intrinsics are identified with Kalibr \cite{furgale2013unified}. ROVIO and MSF employ this calibration data as shown in Fig.~\ref{fig:kalibr}. ${}^{\mathcal{C}}\text{Cam}_{\mbox{\tiny ZR300}}$, ${}^{\mathcal{V}}\text{IMU}_{\mbox{\tiny ZR300}}$, and ${}^{\mathcal{B}}\text{IMU}_{\mbox{\tiny M100}}$ denote an image in $\{\mathcal{C}\}$ frame taken by VI sensor, IMU message in $\{\mathcal{V}\}$ from VI sensor, and IMU message in $\{\mathcal{B}\}$ from Matrice 100's autopilot, respectively. $\boldsymbol{K}$ is an intrinsic camera parameter describing a lens surface using an equidistant distortion model. Note that this procedure only needs to be performed once if the VI sensor configuration (e.g., position and orientation) changes. Otherwise, predefined parameters can be loaded without the calibration process.

\section{Visual-inertial odometry framework}\label{sec:rovio}
\begin{figure}
\center
\includegraphics[width=1\columnwidth]{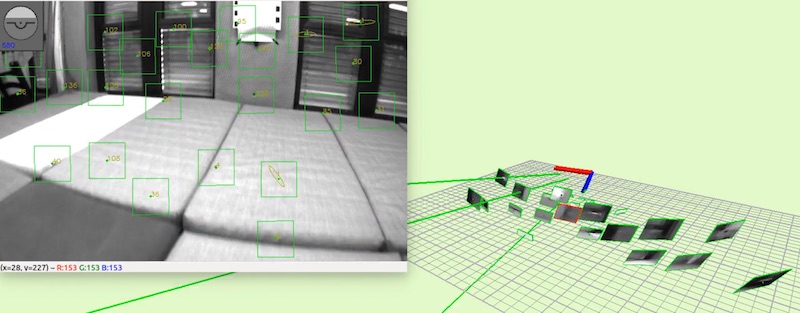}
\caption{A snapshot of ROVIO running with a MAV dataset \cite{burri2016euroc}. The left figure shows tracked landmarks (green dots on the center of green squares and their corresponding IDs). The squares are projected patches which are used for building up multi-level patches from an image pyramid. The predicted landmark location uncertainties are represented by yellow ellipses. The right figure shows 3D estimated camera poses and tracked patches including their distance uncertainties.}
\label{fig:rovio}
\end{figure}

This section introduces our ROVIO framework \cite{Bloesch:2015aa}. The highlights of this approach are three-fold:

\begin{enumerate}

\item It directly leverages pixel intensity errors as an innovation term inside the Extended Kalman Filter (EKF), thereby resulting in a tighter fusion of visual and inertial sensor data. The framework employs multilevel image patches as landmark descriptors (see Fig. \ref{fig:rovio}), therefore computationally expensive the visual feature descriptor extraction step can be avoided. 

\item It makes use of full robocentric formulation to avoid possible corruption of unobservable states. Therefore the consistency of the estimates can be improved. The landmark locations are parametrized by bearing vector and distance parameters with respect to the current camera pose in order to improve modeling accuracy. This is particularly beneficial for fast landmark initialization and circumvents a complicated and cumbersome initialization procedure. The bearing vectors are represented as members of the 2D manifold $S^2$ and minimal differences/derivatives are employed for improved consistency and efficiency (the filter can be run on a single standard CPU core). 

\item The IMU biases and camera-IMU extrinsics are also included into the filter state and co-estimated online for higher accuracy.

\end{enumerate}

Fig. \ref{fig:rovio} illustrates an instance of feature tracking and pose estimation with ROVIO. First, a large number of key point candidates are extracted using a Fast corner detector. Given the current tracked feature set, candidates are filtered out if they are close to the current features based on a threshold (L2-norm distance). After their removal, the adapted Shi-Tomasi score accounting for the combined Hessian on multiple scales is computed for each candidate. A higher Shi-Tomasi score implies better candidate alignment to the corresponding multilevel patch feature. A feature bucketing technique ensures a good distribution of the candidates over the extracted image frame.
An adaptive threshold technique is utilized to regulate the total amount of features (\texttt{25} in this paper) tracked within an EKF based on local (only last a couple of frames) and global (the latest score since it has been detected last time) feature scores. 

It is worth noting that ROVIO requires proper parameter tuning for optimized performance (e.g., the number of features to track, initial covariances, and inverse-depth). Although these parameters must fine-tuned for specific environments, we provide pre-tuned parameters for the camera-IMU (Intel ZR300) and the MAV (Matrice 100) in office-like indoor and farm-site environments.

For EKF modeling and technical detail, we refer the reader to our previous work on the ROVIO framework \cite{Bloesch:2015aa} for further details, including the filter setup, process and measurement models, and their update procedures.


\section{Dynamic systems identification and Non-linear Model Predictive Control}
\label{sec:dynamics}

In this section, we summary our recent work \cite{Sa:2017aa} of MAV dynamic systems identification and describe nonlinear Model Predictive Control (nMPC).
\subsection{Dynamic systems identification}
Our nMPC controller requires first-order attitude (roll and pitch) and thrust dynamics for position control and second-order dynamics for the disturbance observer. To identify these dynamic systems, we record input and output data from manual flights. The inputs are the transmitter commands, roll angle ($u_{\mbox{\tiny $\phi$}}$ in \unit{rad}), pitch angle ($u_{\mbox{\tiny $\theta$}}$ in \unit{rad}), yaw rate ($u_{\mbox{\tiny $\dot{\psi}$}}$ in \unit{rad}/\unit{s}), and thrust ($u_{\mbox{\tiny $\text{z}$}}$ in \unit{N}). The output corresponds to the MAV position, orientation, and linear and angular velocities as provided by a motion capture system. These signals are logged on an onboard computer.

After the data collection, we estimate the scale that maps unit-less transmitter input commands (e.g., -1024$\sim$1024 for pitch command) to the MAV attitude response. This can be determined by linearly mapping with the maximum/minimum angles ($\pm30^\circ$) given maximum/minimum input commands; however, in practice, there can be small errors due to, e.g., an unbalanced platform and subtle dynamic differences. Therefore, these scaling parameters are determined using nonlinear least-squares optimization.

After this process, we use classic system identification techniques given input and output without time delay estimation option and estimated dynamic systems are presented in \cite{Sa:2017aa}.

\def\xref{$\vec{x}^{{\mbox{\tiny ref}}}(t)$}
\def\uref{$\vec{u}^{{\mbox{\tiny ref}}}(t)$}

\subsection{Non-linear MPC for full control}\label{subsec:MPC} 
The MAV controller is based on NMPC \cite{kamelmpc2016}. A cascade approach is used where a high level NMPC generates attitude commands for a low level controller running on the autopilot. The dynamic behavior of the attitude controller is considered as a first order system by the high level controller. The state vector is defined as $ \vec{x} = \left(\vec{p}^{T}, \vec{v}^{T}, \phi, \theta, \psi \right)^{T}$ where $\vec{p}$ and $\vec{v}$ are the MAV position and velocity, respectively, expressed in the inertial frame, \{$\mathcal{W}$\}. We define the control input vector as $\vec{u} = \left(u_{\mbox{\tiny $\phi$}}, u_{\mbox{\tiny $\theta$}}, u_{\mbox{\tiny $T$}} \right)^{T}$, the reference state at time $t$, $\vec{x}^{{\mbox{\tiny ref}}}(t)$, and the steady state control input at time $t$, $\vec{u}^{{\mbox{\tiny ref}}}(t)$. Every time step, the following optimization problem is solved:
\begin{equation} \label{eq:mav_nonlinear_mpc_opt}
\begin{aligned}
\min_{\vec{U}} &\
\int_{t=0}^{T} (\vec{x}(t) - \vec{x}^{{\mbox{\tiny ref}}}(t))^{T}\vec{Q}_{x}(\vec{x}(t) - \vec{x}^{{\mbox{\tiny ref}}}(t)) + (\vec{u}(t) - \vec{u}^{{\mbox{\tiny ref}}}(t))^{T}\vec{R}_{u}(\vec{u}(t) - \vec{u}^{{\mbox{\tiny ref}}}(t)) dt \\ & + (\vec{x}(T) - \vec{x}^{{\mbox{\tiny ref}}}(T))^{T}\vec{P}(\vec{x}(T) - \vec{x}^{{\mbox{\tiny ref}}}(T))\textrm{,}\\
&\begin{aligned}
\text{subject to} &
& & \dot{\vec{x}} = \vec{f}(\vec{x}, \vec{u})\textrm{;}\\
& & & \vec{u}(t) \in \mathcal{U}_{C}\textrm{,} \\
& & & \vec{x}(0) = \vec{x}\left( {t_0}\right)\textrm{,}
\end{aligned}
\end{aligned}
\end{equation}

\noindent where $ \vec{Q}_{x} \succeq 0 $ is the penalty on the state error, $ \vec{R}_{u} \succ 0 $ is the penalty on control input error, and $ \vec{P} $ is the terminal state error penalty. The $\succeq$ operator denotes the positive definiteness of a matrix. $ \vec{f}(\vec{x}, \vec{u}) $ is the ordinary differential equation representing the MAV dynamics. This term is given by:
\begin{subequations}
	\begin{eqnarray}
	\dot{\vec{p}} &=& \vec{v} ,\label{eq:dynamics_eq1} \\
	\dot{\vec{v}} &=& \frac{1}{m}\left( \vec{R}_{\mathcal{W}, \mathcal{B}}\left[
	\begin{array}{ccc}
	0\\
	0\\
	u_{\mbox{\tiny $T$}}
	\end{array}
	\right] - u_{\mbox{\tiny $T$}}\vec{K}_{\mbox{\tiny drag}}\vec{v} + \vec{F}_{{\mbox{\tiny ext}}}\right) + \left[ \begin{array}{ccc}
	0 \\ 
	0 \\ 
	-g
	\end{array}  \right], \label{eq:dynamics_eq2} \\
    \dot{\phi} &=& \frac{1}{\tau}_{\phi} \left(k_{\phi} u_{\mbox{\tiny $\phi$}} - \phi  \right), \label{eq:roll_dynamics} \\ 
	\dot{\theta} &=& \frac{1}{\tau}_{\theta} \left( k_{\theta}u_{\mbox{\tiny $\theta$}} - \theta \right),\label{eq:pitch_dynamics}\\ 
	\dot{\psi} &=& u_{\mbox{\tiny $\dot{\psi}$}},  \label{eq:yaw_dynamics}
	\end{eqnarray}
\end{subequations}
\noindent where $ m $ is the vehicle mass, $ \vec{R}_{\mathcal{W}, \mathcal{B}} $ is the rotation matrix from body frame $\mathcal{B}$ to inertial frame $\mathcal{W}$, $\vec{K}_{\mbox{\tiny drag}}=diag(k_{\mbox{\tiny d}}, k_{\mbox{\tiny d}}, 0)$ is the drag coefficient matrix, $ \vec{F}_{\mbox{\tiny ext}} $ are the external forces acting on the vehicle (such as wind gusts), $g$ is gravitational acceleration, and $\phi, \theta, \psi$ represent the roll, pitch and yaw angles of the vehicle. $\tau_{\phi}, \tau_{\theta}$ are the time constants of the roll and pitch dynamics, respectively. $k_{\phi}, k_{\theta}$ are the gains of the roll and pitch dynamics, respectively. $ u_{\mbox{\tiny $\dot{\psi}$}}$ is the heading angular rate command. Note that we assume perfect tracking of the heading angular rate as it does not affect the vehicle position. $ \vec{F}_{\mbox{\tiny ext}} $ is estimated in real-time using an augmented Kalman Filter as described in  \cite{kamelmpc2016}.


\section{Experimental results}\label{sec:results}
We present our implementation details, hardware and software setup, and evaluate control and state estimation performance in indoor and outdoor environments. We perform 9 experiments in different conditions while varying tasks to demonstrate the repeatability and feasibility of the proposed approach. Tables \ref{tbl:ctrl_results_summary} and \ref{tbl:esti_results_summary} summarize the control and state estimation performances as root-mean-square (RMS) error. A detailed result analysis is presented in the following sections.

\begin{table}[]
\centering
\caption{Control performance (RMS error) summary}
\label{tbl:ctrl_results_summary}
\begin{tabular}{ccccccccccc}
\hline
& \multicolumn{3}{c}{\textbf{Hovering}} & \multicolumn{3}{c}{\textbf{Step response}} & \multicolumn{3}{c}{\textbf{Trj. following}} &Unit      \\ \hline
& \multicolumn{2}{c}{Indoor} &Outdoor & \multicolumn{2}{c}{Indoor} &Outdoor & \multicolumn{2}{c}{Indoor} &Outdoor &      \\ \hline
Pose                   &                              \textbf{0.036}&    0.049  & 0.045                         & 0.233                    &     0.395      & 0.260           &    0.083                    &       0.091     & 0.100      & \unit{m}    \\ \hline
x                        &                              \textbf{0.016}&         0.022      & 0.030                &  0.155                   &      0.277      & 0.189          &       0.066                 &        0.042       & 0.079         & \unit{m}    \\ \hline
y                        &                    0.018           &          \textbf{0.012}      & 0.026               &   0.125                  &      0.230        & 0.172        &      0.039                  &        0.071      & 0.056           & \unit{m}    \\ \hline
z                        &                      0.026         &          0.038       & 0.022              &   0.122                  &     0.163      & 0.049           &     0.030                   &        0.038    & \textbf{0.024}          & \unit{m}    \\ \hline
roll                     &                       \textbf{0.863}        &            1.389       & ---            &   ---                  &         ---     & ---        &       1.396               &   1.593    & ---            & \unit{deg}  \\ \hline
pitch                    &                      \textbf{0.793}         &          0.913        & ---             &  ---                   &       ---     & ---          &       0.871                &  1.067    & ---                & \unit{deg}  \\ \hline
yaw                      &                   1.573            &          3.024       & ---              &  3.659                   &     6.865      & ---           &        \textbf{1.344}               &      2.858    & ---      & \unit{deg}  \\ \hline
Duration                 &           30-230                    &        50-180      & 20-150                &   30-200                  &   20-120     & 20-120              &   30-80                     &    25-120      & 50-180             & \unit{s}    \\ \hline
Wind               & ---                           &         11-11.5   & 3.6-7.4     &   ---               &       11-11.5    & 3.6-7.4           & ---                    &     11-11.5     & 3.6-7.4         & \unit{m/s}  \\ \hline
\end{tabular}
\end{table}

\begin{table}[]
\centering
\caption{State estimation performance (RMS error) summary}
\label{tbl:esti_results_summary}
\begin{tabular}{ccccccccccc}
\hline
& \multicolumn{3}{c}{\textbf{Hovering}} & \multicolumn{3}{c}{\textbf{Step response}} & \multicolumn{3}{c}{\textbf{Trj. following}} &Unit      \\ \hline
& \multicolumn{2}{c}{Indoor} &Outdoor & \multicolumn{2}{c}{Indoor} &Outdoor & \multicolumn{2}{c}{Indoor} &Outdoor &      \\ \hline
Pose                   &                              \textbf{0.013} &    0.019  & 0.043                         & 0.118                    &     0.133      & 0.091           &    0.091                    &       0.097    & 0.099      & \unit{m}    \\ \hline
x                        &                              \textbf{0.008} &         0.010      & 0.026                &  0.103                   &      0.107      & 0.054          &       0.052                 &        0.075       & 0.084         & \unit{m}    \\ \hline
y                        &                    \textbf{0.008}           &          0.014      & 0.028               &   0.028                  &      0.048        & 0.062        &      0.062                  &        0.078      & 0.048           & \unit{m}    \\ \hline
z                        &                      \textbf{0.007}         &          \textbf{0.007}      & 0.019              &   0.050                  &     0.062      & 0.041           &     0.042                   &        0.065    & 0.022          & \unit{m}    \\ \hline
roll                     &                       \textbf{0.160}        &            0.322       & ---            &   ---                  &         ---     & ---        &       0.857               &   0.345    & ---            & \unit{deg}  \\ \hline
pitch                    &                      \textbf{0.103}         &          0.291        & ---             &  ---                   &       ---     & ---          &       0.911                &  0.309    & ---                & \unit{deg}  \\ \hline
yaw                      &                   \textbf{0.813}            &          0.977       & ---              &  2.704                   &     3.789      & ---           &        0.883                &      0.907    & ---      & \unit{deg}  \\ \hline
Duration                 &           30-230                    &        50-180      & 20-150                &   30-200                  &   20-120     & 20-120              &   30-80                     &    25-120      & 50-180             & \unit{s}    \\ \hline
Wind               & ---                           &         11-11.5   & 3.6-7.4     &   ---               &       11-11.5    & 3.6-7.4           & ---                    &     11-11.5     & 3.6-7.4         & \unit{m/s}  \\ \hline
\end{tabular}
\end{table}

\subsection{Hardware Setup}
Our quadcopter MAV carries an Intel NUC 5i7RYH (i7-5557U, 3.1\unit{GHz} dual cores, 16\unit{GB} RAM), running Ubuntu Linux 14.04 and ROS Indigo onboard \cite{Quigley:2009aa}. It is equipped with a flight controller, N1 autopilot, an embedded IMU providing vehicle orientation, acceleration, and angular velocity at 50\unit{Hz} to the computer via 921,600\unit{bps} USB-to-serial communication. We mount a down-facing VI sensor (the distance from $\mathcal{C}$ to the ground is 0.106\unit{m} sitting on a flat surface) and activate only a fisheye camera at 30\unit{Hz} and IMU at 200\unit{Hz} (disabling the stereo IR cameras and IR projector).

The total system mass is 3.62\unit{kg} and the MAV carries 1.27\unit{kg} payload including the onboard computer, a gimbal camera, and a VI sensor. A 6-cell LiPo battery (22.2\unit{V}, 4500\unit{mAh}) powers the device with total flight time of $\approx12\unit{mins}$ without any computational load (only running the autopilot with a small angle of attack $\approx\pm20^{\circ}$) and $\approx10\unit{mins}$ 50\unit{s} with all process running\footnote{Low-battery threshold is set \%20 of the battery.}. The MAV and a ground station are connected via WiFi with proper time synchronization.

\subsection{Software setup}

\begin{figure}
\centering
	\subfloat[]{\includegraphics[height=0.31\columnwidth]{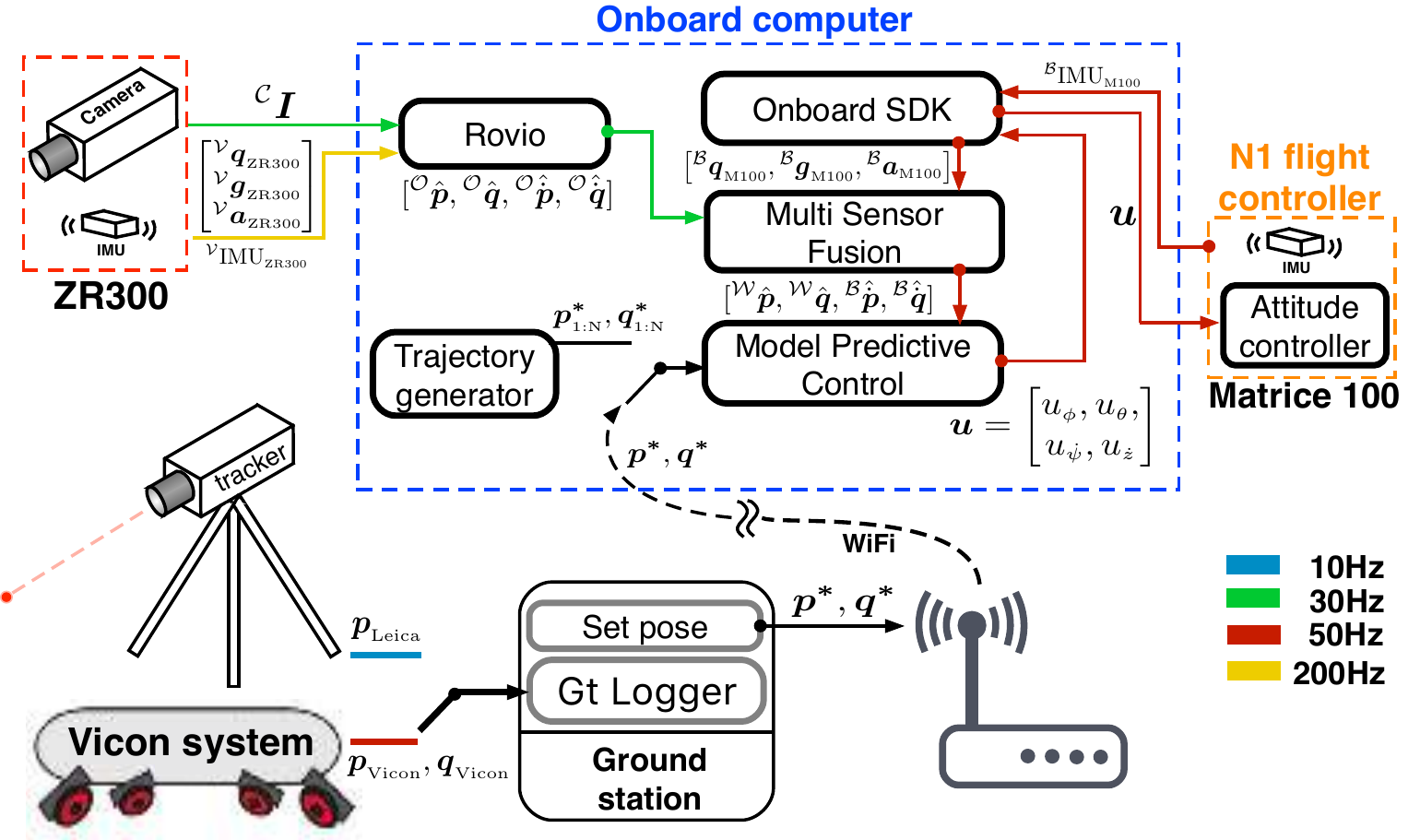}}
	\subfloat[]{\includegraphics[height=0.31\columnwidth]{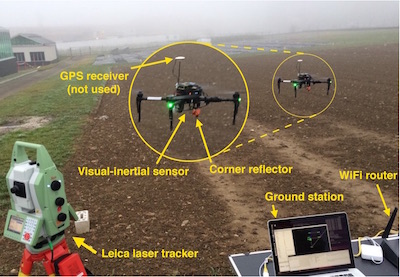}}
\caption{ (a) illustrates our software architecture, with different colors denoting the corresponding sample rates. (b) shows our outdoor experimental setup. }
\label{fig:diagram}
\end{figure}

Our system is integrated using ROS as shown in Fig.~\ref{fig:diagram}. Each box represents a ROS node running at different rates. Rovio receives the fisheye images ${}^{\mathcal{C}}\boldsymbol{I}$ and IMU measurements $\begin{bmatrix}
{}^{\mathcal{V}}\boldsymbol{q}_{\mbox{\tiny ZR300}},
{}^{\mathcal{V}}\boldsymbol{g}_{\mbox{\tiny ZR300}},
{}^{\mathcal{V}}\boldsymbol{a}_{\mbox{\tiny ZR300}}
\end{bmatrix}$. The estimated odometry $\begin{bmatrix}
{}^{\mathcal{O}}\hat{\boldsymbol{p}},{}^{\mathcal{O}}\hat{\boldsymbol{q}},{}^{\mathcal{O}}\hat{\dot{\boldsymbol{p}}},{}^{\mathcal{O}}\hat{\dot{\boldsymbol{q}}}
\end{bmatrix}$ is subscribed by the MSF framework \cite{Weiss:2011aa} to increase the rate from 30\unit{Hz} to 50\unit{Hz} using IMU from N1 flight controller $\begin{bmatrix}
{}^{\mathcal{B}}\boldsymbol{q}_{\mbox{\tiny M100}},
{}^{\mathcal{B}}\boldsymbol{g}_{\mbox{\tiny M100}},
{}^{\mathcal{B}}\boldsymbol{a}_{\mbox{\tiny M100}}
\end{bmatrix}$. The output from MSF $\begin{bmatrix}
{}^{\mathcal{W}}\hat{\boldsymbol{p}},{}^{\mathcal{W}}\hat{\boldsymbol{q}},{}^{\mathcal{B}}\hat{\dot{\boldsymbol{p}}},{}^{\mathcal{B}}\hat{\dot{\boldsymbol{q}}}
\end{bmatrix}$ is fed to the subsequent MPC position controller. Finally, the control output $\boldsymbol{u}=\begin{bmatrix}
u_{\mbox{\tiny $\phi$}}, u_{\mbox{\tiny $\theta$}},u_{\mbox{\tiny $\dot{\psi}$}},u_{\mbox{\tiny $\dot{z}$}}
\end{bmatrix}$ are transmitted to the attitude flight controller.

The ground station sets either a goal pose $[\boldsymbol{p^{*}},\boldsymbol{q^{*}}]$ for position and orientation or N sequences $[\boldsymbol{p_{\mbox{\tiny 1:N}}^{*}},\boldsymbol{q_{\mbox{\tiny 1:N}}^{*}}]$ created by the onboard trajectory generator \cite{burri2016maximum}.

For the indoor and outdoor experiments, we utilize the Vicon motion capture system and the Leica laser tracker, respectively, to obtain ground truth. Note that the Vicon system can provide position and orientation $[\boldsymbol{p}_{\mbox{\tiny Vicon}},\boldsymbol{q}_{\mbox{\tiny Vicon}}]$ whereas the laser tracker can only provide position ground truth $\boldsymbol{p}_{\mbox{\tiny Leica}}$.

It is important to properly tune the MSF parameters, such as measurement and process noise. This particularly impacts vertical state estimation (altitude and vertical velocity) since the VI sensor is downwards-facing. With our setting, the MAV can fly up 15\unit{m} without experiencing issues.

\subsection{Experimental setup}
We perform 9 experiments in both indoor and outdoor environments to evaluate control and state estimation with respect to ground truth. More specifically, 3 tasks are considered: hovering, step response, and trajectory following. To demonstrate controller robustness, wind disturbances are generated in the indoor environment using a fan with 260\unit{W} and 300\unit{m^3/min} air flow. As measured by an anemometer, this produces a 11-11.5\unit{m/s} disturbance in the hovering position.

To evaluate control performance indoors, we compute the root-mean-square (RMS) error between the reference and actual vehicle positions and orientations as obtained from the motion capture device, with Euclidean distance used for calculating the differences between 3D poses. Outdoors, we only consider positional error with respect to ground truth from the laser tracker. Similarly, state estimation performance is quantified using RMS error between ground truth and pose estimates.


\subsection{Performance evaluation}
We evaluate control and state estimation performance evaluation using RMS error for 9 experiments. For control and state estimation performance, RMS error between reference commands and ground truth poses is computed for the former and the error between the ground truth and estimated pose is used for the latter. Qualitative results are also demonstrated for short-long trajectory following tasks. Due to space limitations, only a subset of result plots is presented while Tables \ref{tbl:ctrl_results_summary} and \ref{tbl:esti_results_summary} provide a complete result summary.

\subsubsection{Control performance evaluation}
\begin{figure}
\centering
\subfloat[]{\includegraphics[width=0.49\columnwidth]{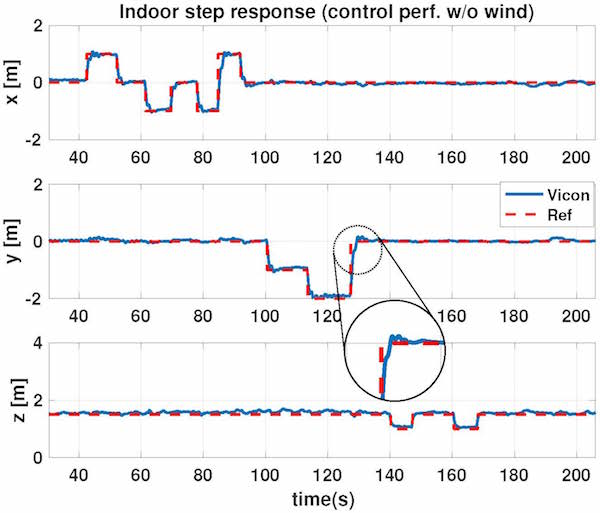}}
\subfloat[]{\includegraphics[width=0.49\columnwidth]{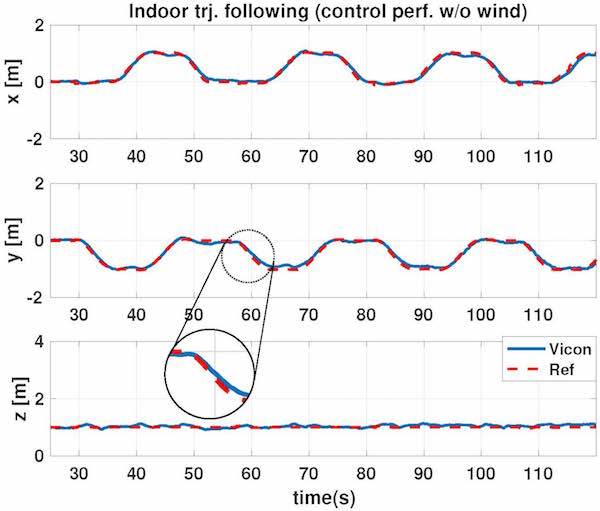}}\newline
\subfloat[]{\includegraphics[width=0.49\columnwidth]{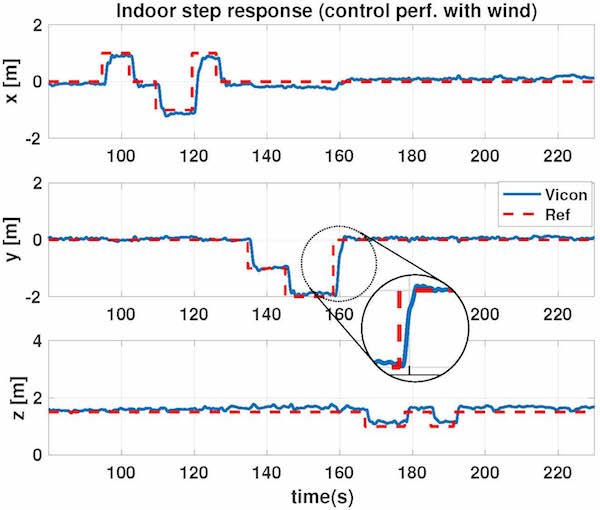}}
\subfloat[]{\includegraphics[width=0.49\columnwidth]{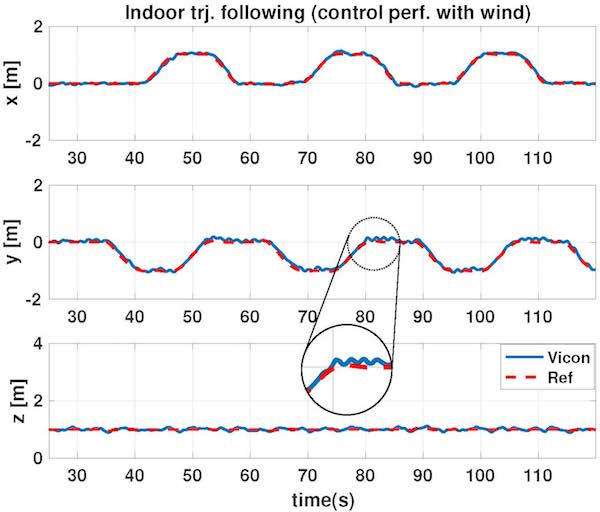}}\newline
\subfloat[]{\includegraphics[width=0.49\columnwidth]{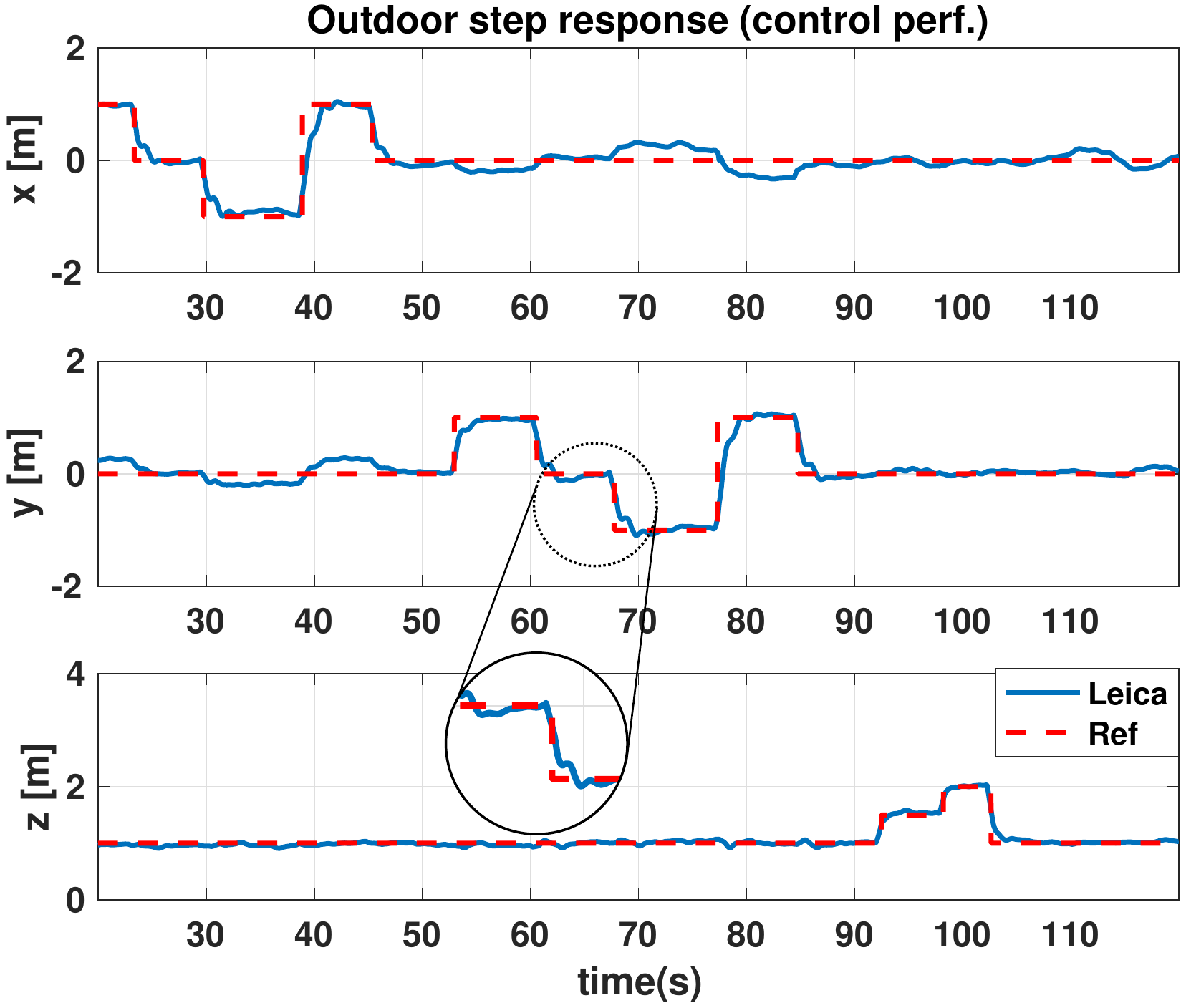}}
\subfloat[]{\includegraphics[width=0.49\columnwidth]{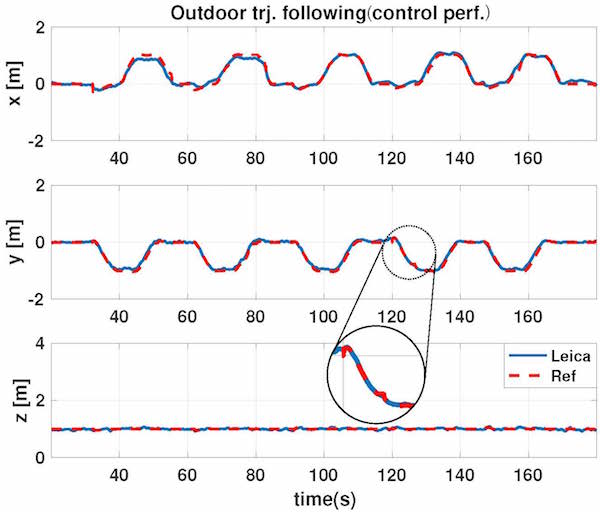}}

\caption{Step response (left column) and trajectory following (right column) control performance in indoor (a)$\sim$(d) and outdoor environments, (e) and (f).}
\label{fig:ctrl_perf}
\end{figure}

Despite their resistance to external disturbances, the downside of using heavy platforms is their slower response. Fig.~\ref{fig:ctrl_perf} shows step response plots (a) without and (c) with wind disturbances in indoor and (e) outdoor environments (first column). A goal position is manually chosen to excite all axes. The results in Table~\ref{tbl:ctrl_results_summary} evidence a relatively large control error in both the x- and y-directions, because slow response causes errors to accumulate as the goal reference is reached. Moreover, in Fig.~\ref{fig:wp}, we use the method of Burri et al. \cite{burri2015real} to generate a smooth polynomial reference trajectory.
Fig.~\ref{fig:ctrl_perf} (b), (d), and (f) show gentle trajectory following control performance results, and (g) and (h) display more aggressive and agile trajectory following performance. The trajectory is configured as 10.24\unit{m} with maximum velocity and acceleration of 1.63\unit{m/s}, and 5.37\unit{m/s^2} given 9.07\unit{s} time budget \cite{bahnemann2017sampling}. Note that yaw tracking error is quite large because of physical hardware limitations (i.e., M100's maximum angular rate given the trajectory and payload). The accuracy of trajectory following in comparison to hovering implies that the proposed approach is applicable in many practical applications, such as obstacle avoidance or path planning. 

\begin{Contfigure}
\ContinuedFloat
\centering
\subfloat[]{\includegraphics[width=0.49\columnwidth]{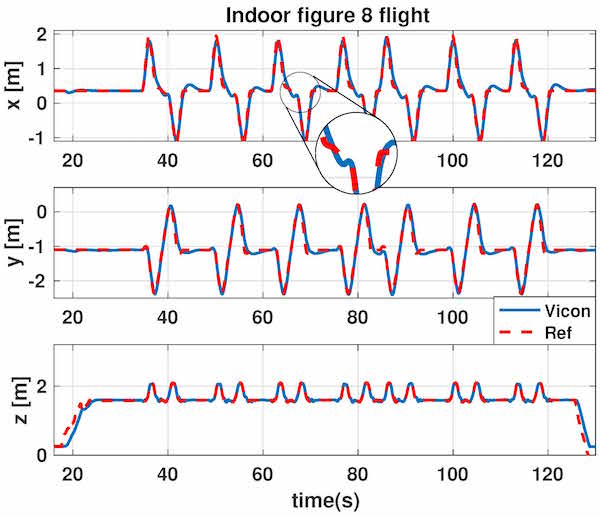}}
\subfloat[]{\includegraphics[width=0.49\columnwidth]{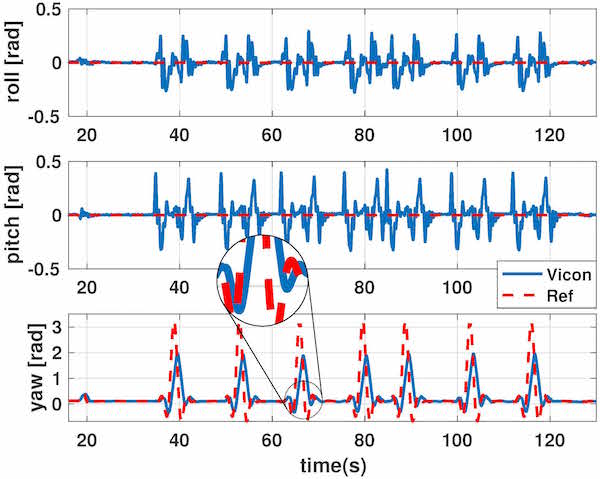}}

\caption{(g) and (h) show position and orientation control performance while the MAV tracks an aggressive trajectory 7 times.}
    \label{fig:ctrl_perf_cont}
\end{Contfigure}


\subsubsection{State estimation performance evaluation}

\begin{figure}
\centering
\subfloat[]{\includegraphics[width=0.49\columnwidth]{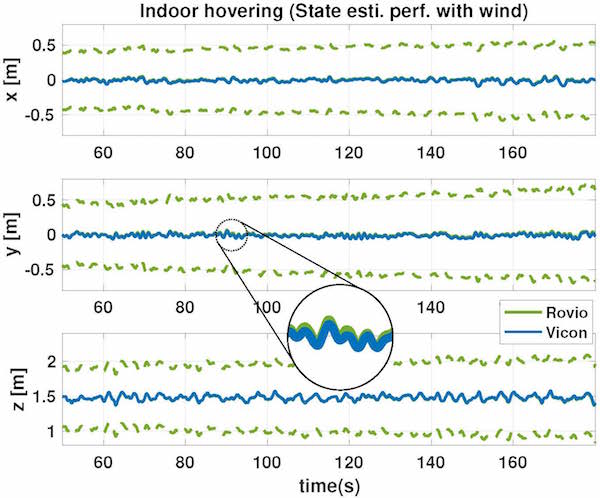}}
\subfloat[]{\includegraphics[width=0.49\columnwidth]{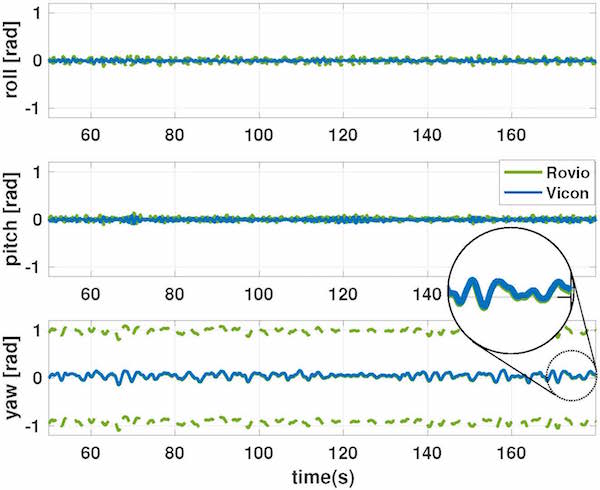}}\newline
\subfloat[]{\includegraphics[width=0.49\columnwidth]{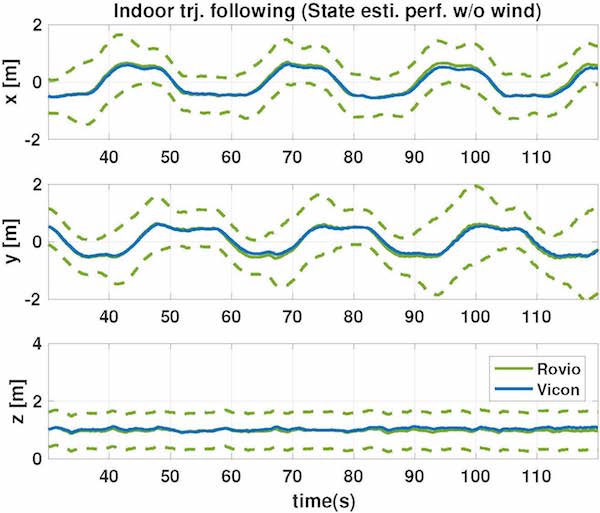}}
\subfloat[]{\includegraphics[width=0.49\columnwidth]{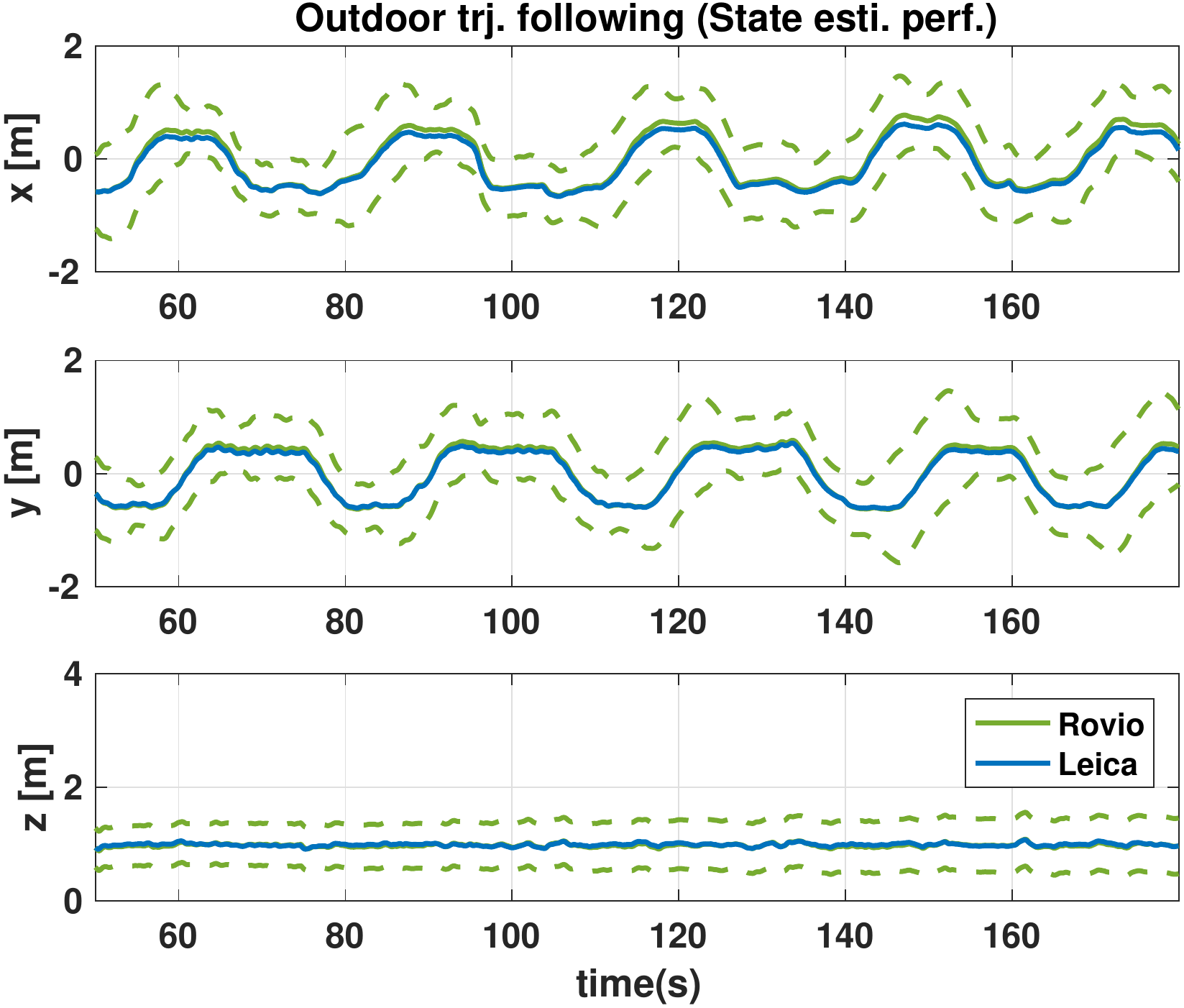}}\newline
\subfloat[]{\includegraphics[width=0.49\columnwidth]{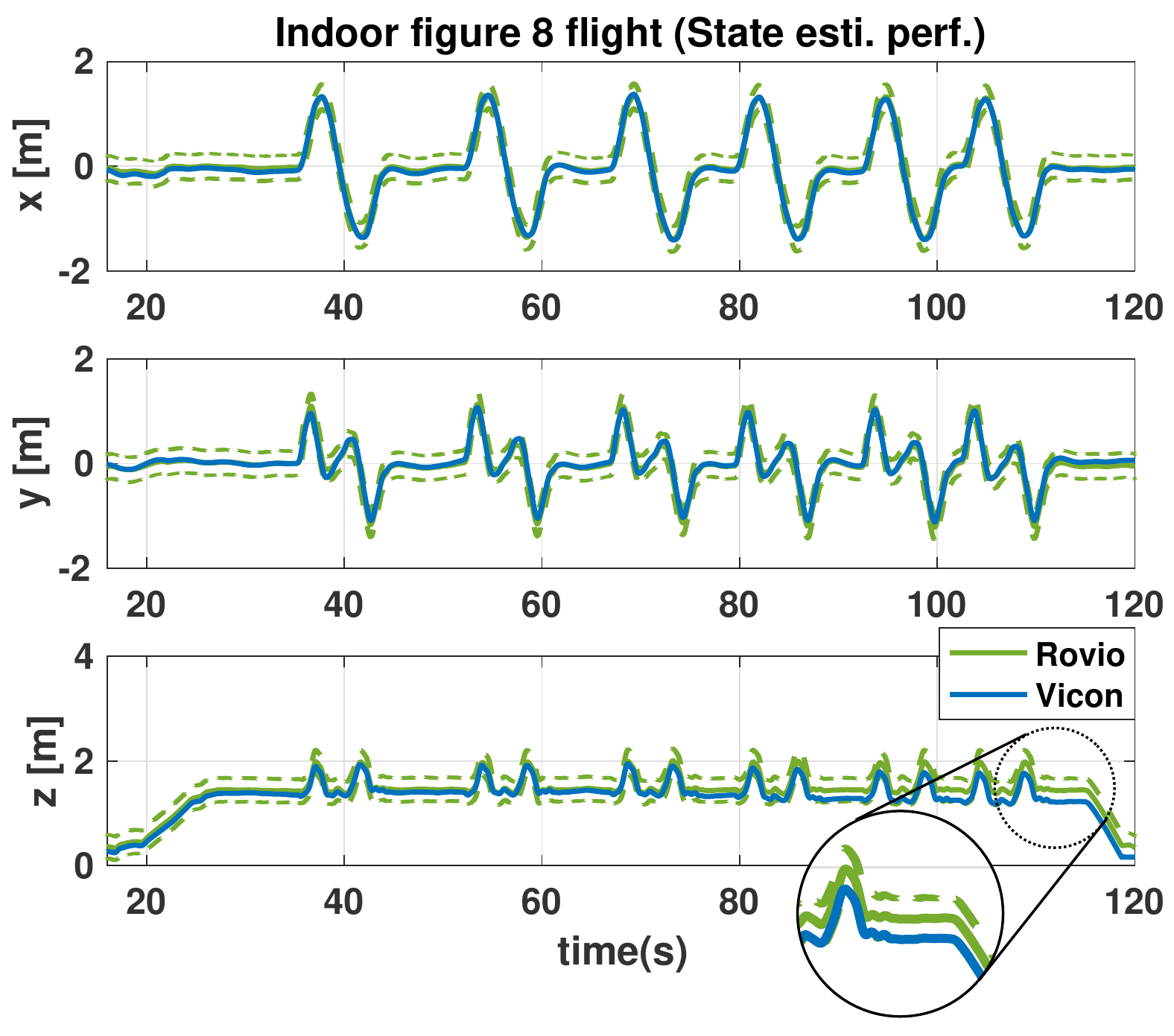}}
\subfloat[]{\includegraphics[width=0.51\columnwidth]{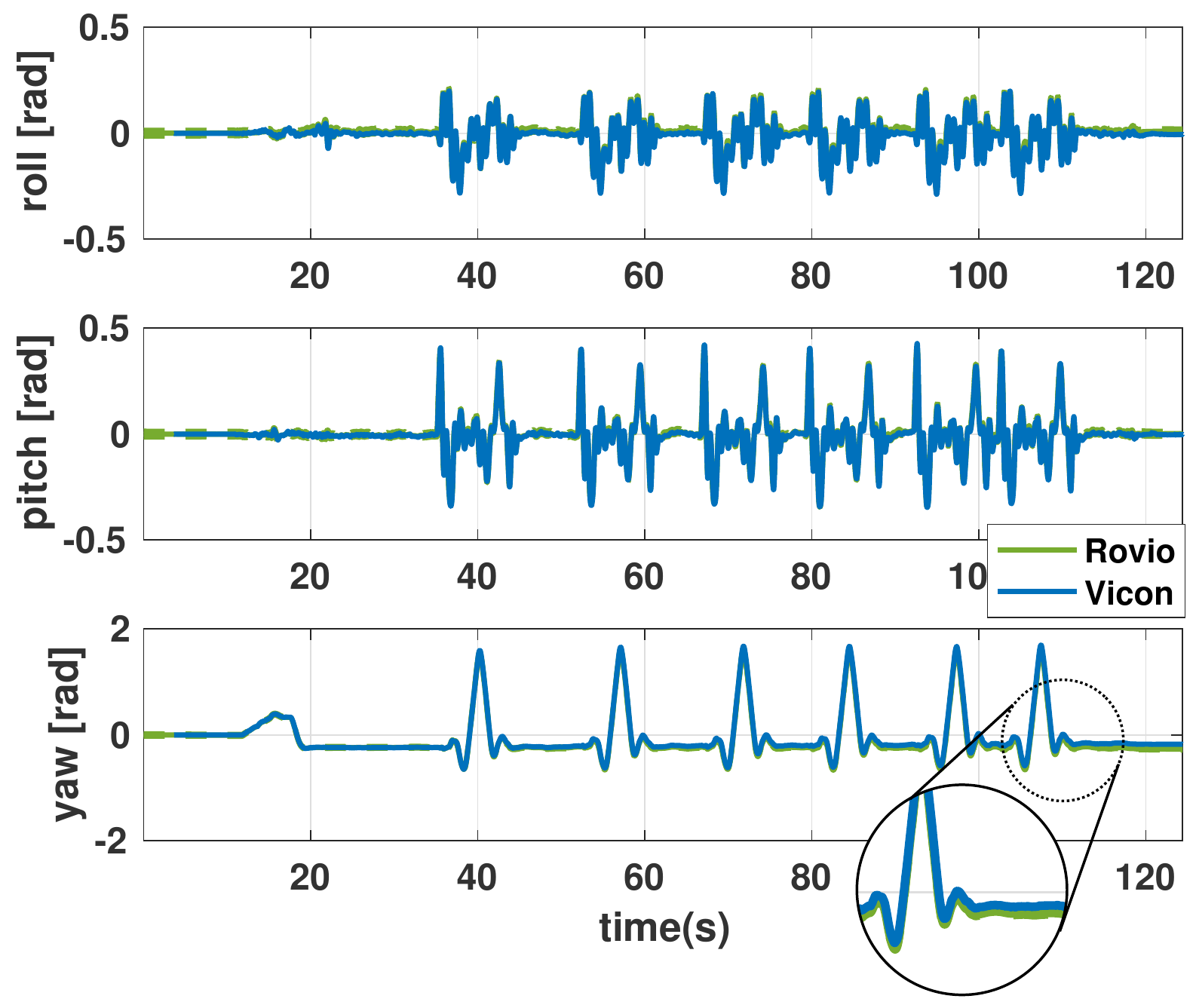}}
\centering
\caption{(a) position and (b) orientation state estimation in hover, and position estimation while performing trajectory following (c) indoors and (d) outdoors. (e) and (f) are position and orientation state estimation respectively during aggressive trajectory following.}
\label{fig:esti_result}
\end{figure}

Fig.~\ref{fig:esti_result} (a) and (b) depict position and orientation estimation using Rovio and the VI sensor while hovering with wind disturbances. The plots illustrate the disturbances continuously pushing the MAV, incurring control error, whereas estimation drifts very slowly within the $\pm3\sigma$ boundary. Fig.~\ref{fig:esti_result} (c) and (d) show position estimation indoors and outdoors while performing trajectory following. Note that the performance can vary slightly depending on the flying environment due to visual feature differences. Table \ref{tbl:ctrl_results_summary} shows that variations in state estimation between tasks are smaller than those of control performance. This implies the control performance error can be caused by physical vehicle limitations (e.g., motor saturation), imperfect dynamics modeling, and manual controller tuning. The last two figures (e) and (f) show accurate state estimation while tracking aggressive figure of 8 trajectories with varying height and yaw 6 times. Almost states lie within the $\pm3\sigma$ boundary but it can be clearly seen that height and yaw estimation drift around 110\unit{s} due to accumulated errors caused by fast maneuvers.

\subsubsection{Qualitative results}
\begin{figure}
\centering
\subfloat[Indoor w/o disturbances.]{\includegraphics[width=0.32\columnwidth]{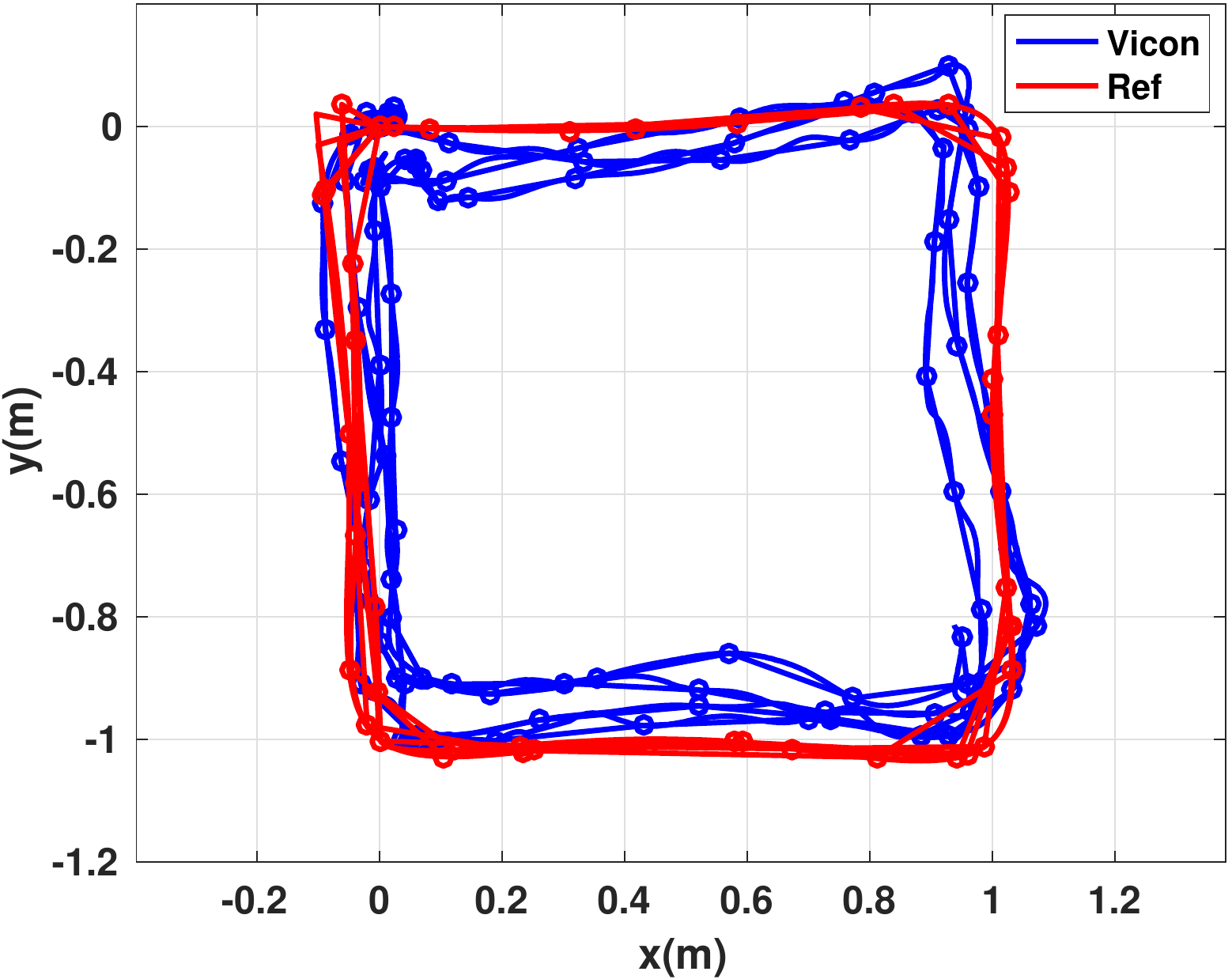}}
\subfloat[Indoor with disturbances.]{\includegraphics[width=0.32\columnwidth]{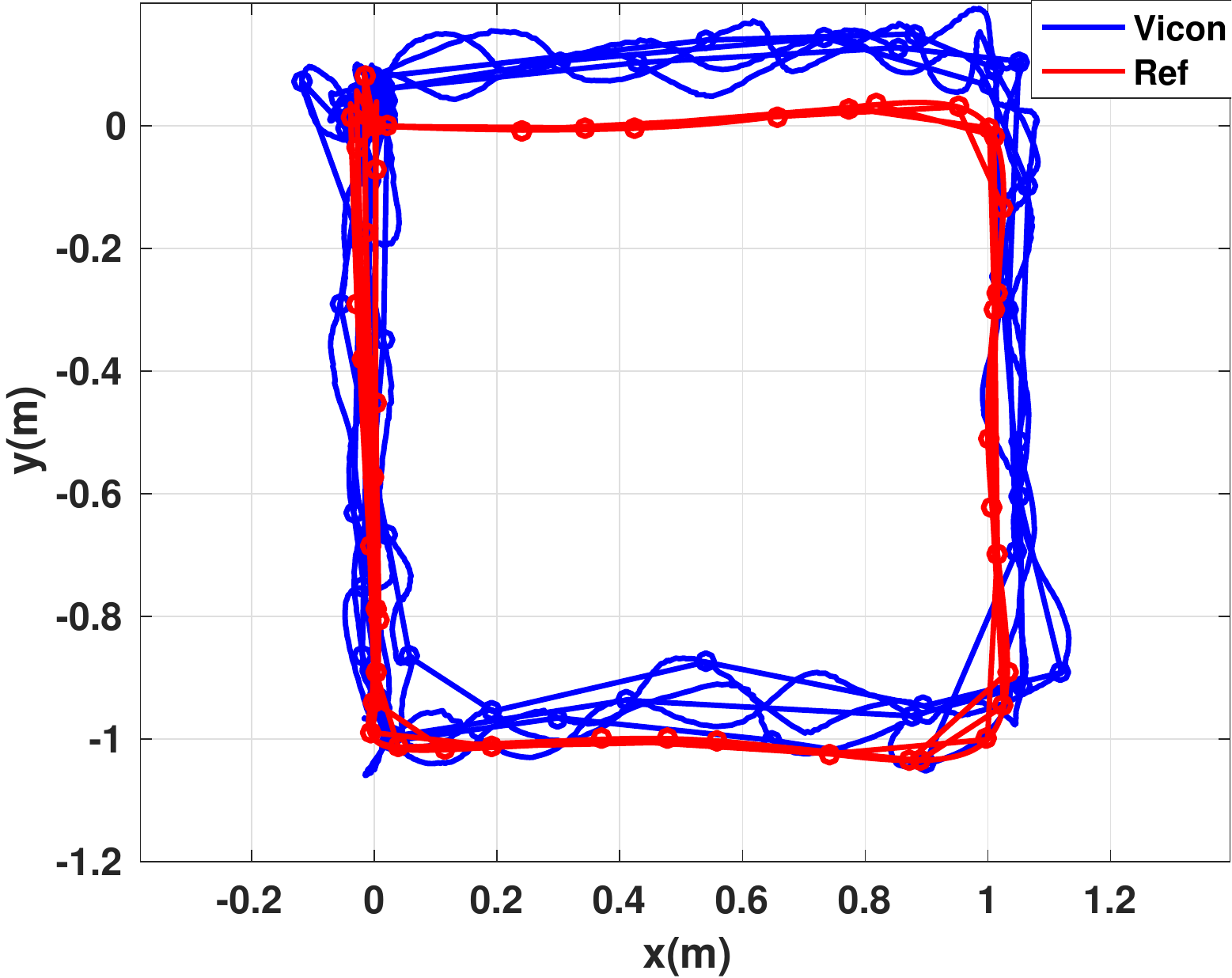}}
\subfloat[Outdoor]{\includegraphics[width=0.32\columnwidth]{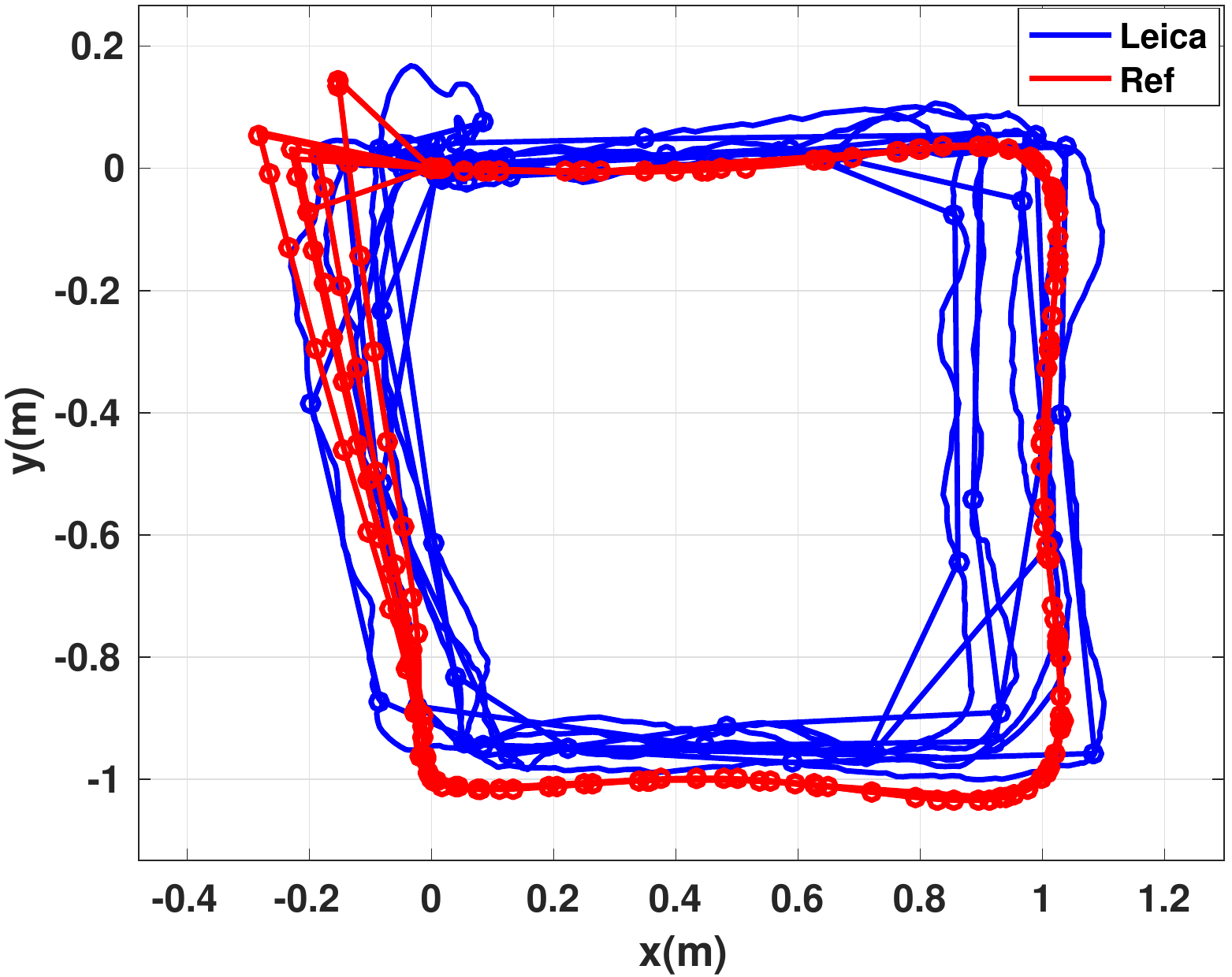}}\newline
\subfloat[Indoor w/o disturbances.]{\includegraphics[width=0.32\columnwidth]{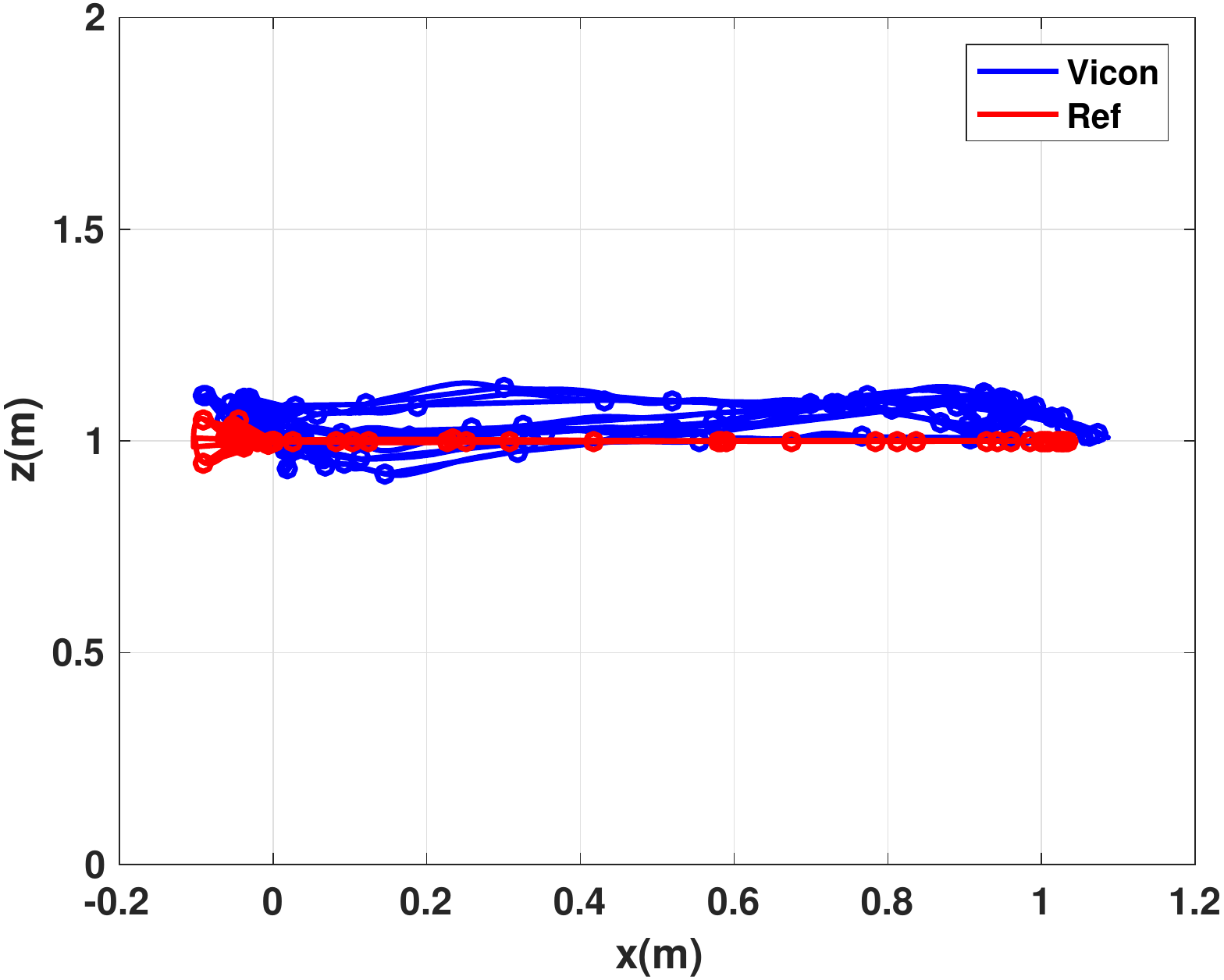}}
\subfloat[Indoor with disturbances.]{\includegraphics[width=0.32\columnwidth]{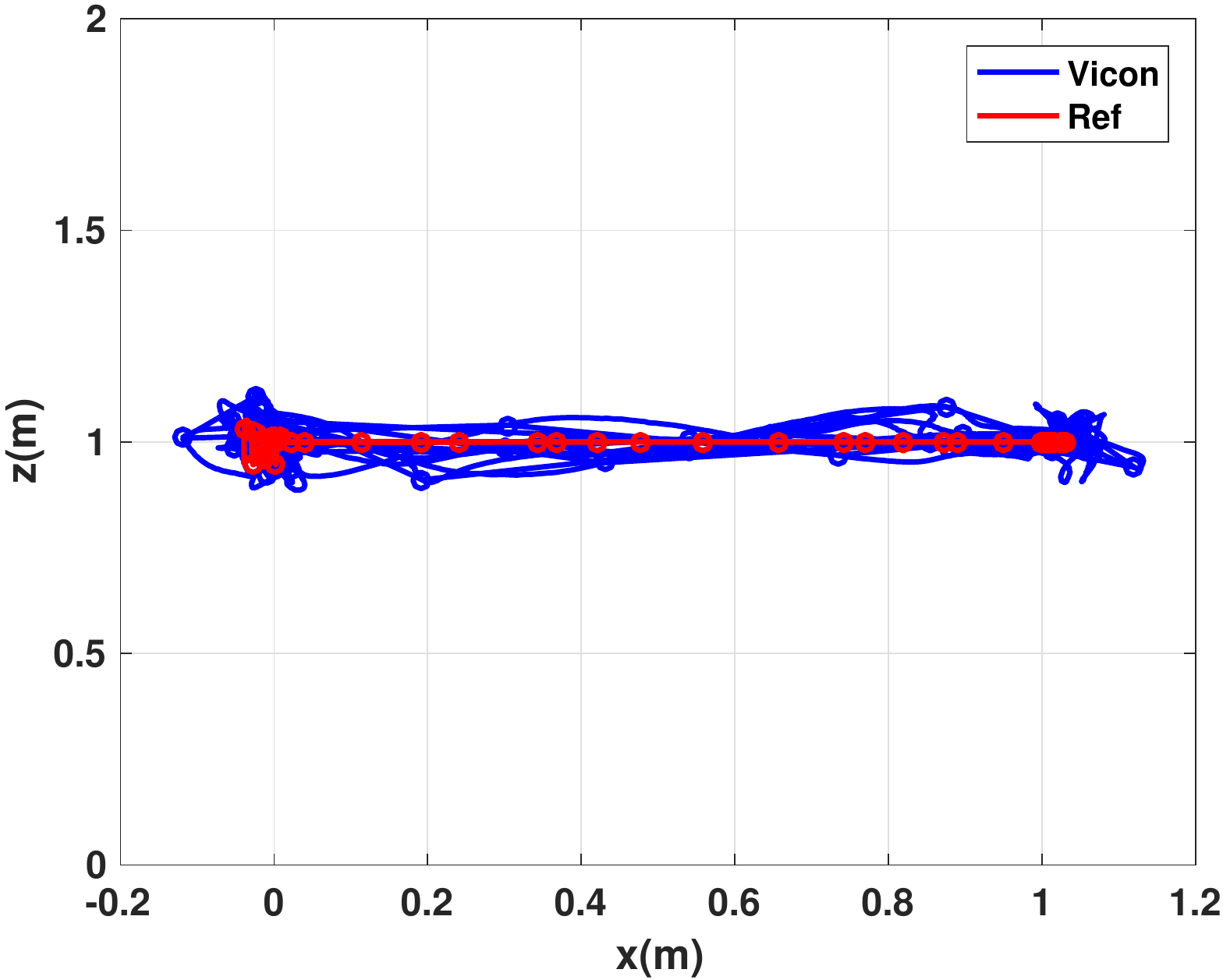}}
\subfloat[Outdoor]{\includegraphics[width=0.32\columnwidth]{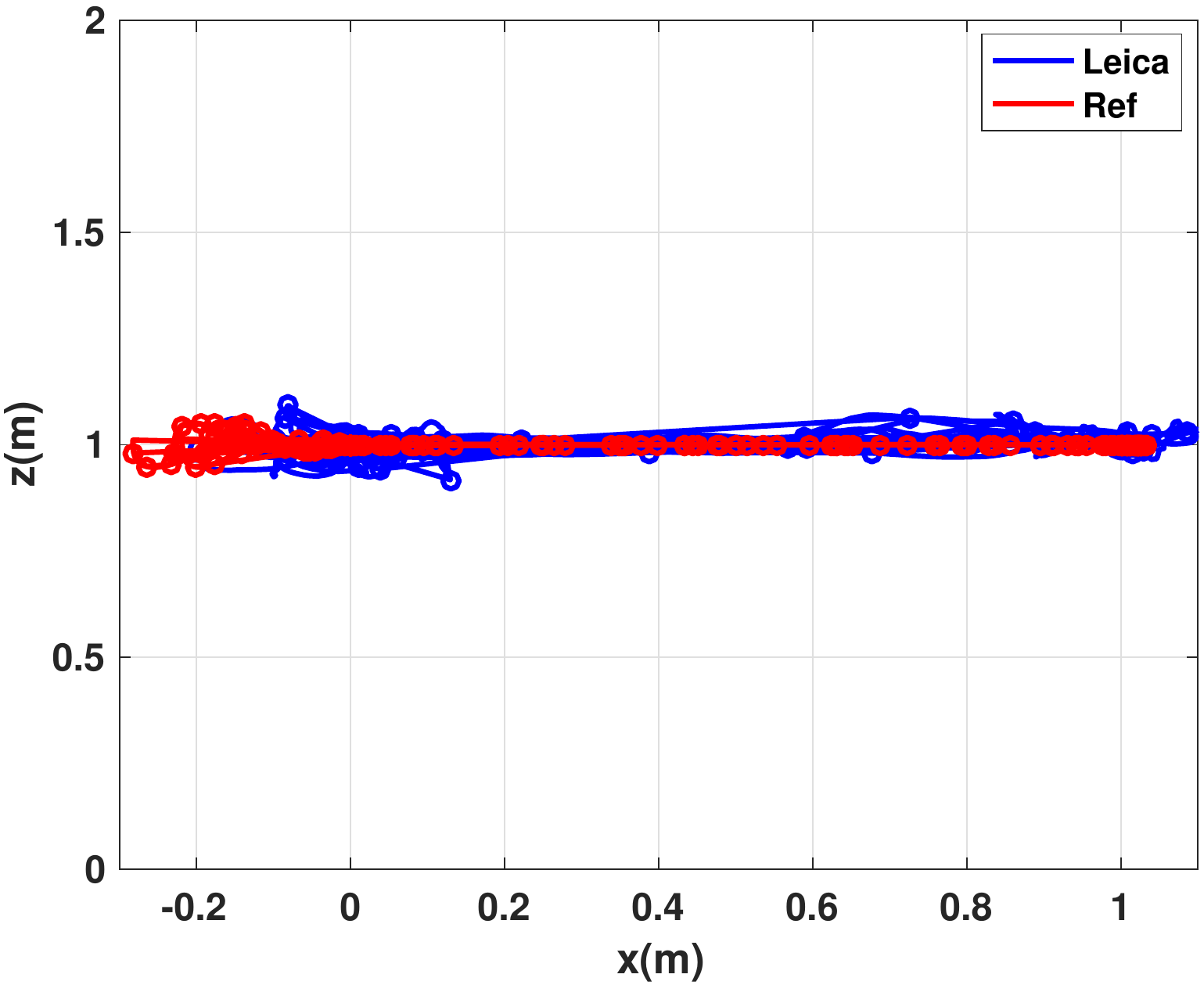}}\newline
\subfloat[]{\includegraphics[width=0.6\columnwidth]{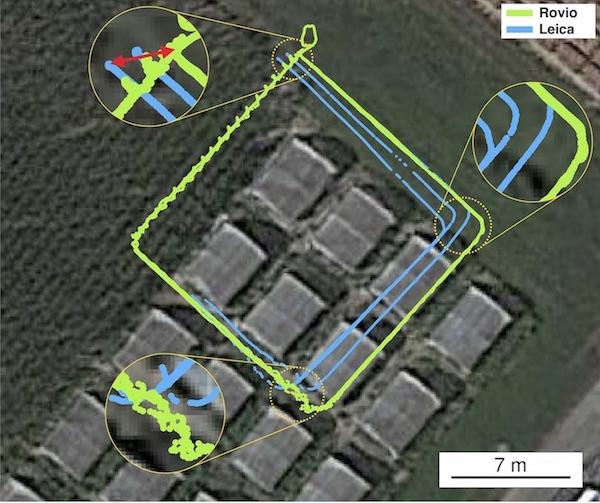}}
\centering
\vspace{-3mm}
\caption{Qualitative results for trajectory following indoors and outdoors. A fan around 3\unit{m} away from the MAV was used to introduce external wind disturbances. The figure in (g) shows a longer distance flight ($\approx180\unit{m})$ in outdoor farm site. Note that the laser tracker lost track mid-flight due to occlusions by the vehicle itself. The red arrow depicts take-off and landing positions where a qualitative drift error is calculated.}
\label{fig:wp}
\vspace{-5mm}
\end{figure}

Fig.~\ref{fig:wp} presents two qualitative results for short and long trajectory following. The first and second row are top and side views, respectively. Red illustrates the planned trajectory and the MAV position is marked in blue. Note that a fan is located around 3\unit{m} away from the origin (i.e., x\,=\,0, y\,=\,0) along the South-East direction. The trajectory shift due to wind in the positive x- and -y-directions is evident. The results for a long trajectory following task are depicted in Fig.~\ref{fig:wp} (g). The length of one side of the square is around 15\unit{m} and the vehicle flies around the area along the side 3 times ($\approx180\unit{m}$). The plot shows that visual odometry drifts while flying at 1.288\unit{m}, with the red arrow marking the offset between taking-off and landing positions. Qualitatively, the 0.82\% error of the total flight distance is consistent with our previous results \cite{Bloesch:2015aa}.

\section{Conclusions}\label{sec:conclusion}
We have presented state estimation and control performance of a VI-aided cost-effective VTOL MAV platform. The combination of robust visual odometry and a state-of-the-art controller with traditional dynamic system identification brings commercial products (MAVs and visual-inertial sensors) into the research domain. The applied methods were evaluated in both indoor and outdoor environments. Competitive results demonstrate that our approach represents a stepping stone towards achieving more sophisticated tasks. We return our experiences and findings to the community through open-source documentation and software packages to support researchers building custom VI-aided MAVs.
\subsubsection*{Acknowledgement}
This project has received funding from the European Union's Horizon 2020 research and innovation programme under grant agreement No 644227 and No 644128 from the Swiss State Secretariat for Education, Research and Innovation (SERI) under contract number 15.0029 and 15.0044.

\bibliographystyle{ieeetr}
\bibliography{RAM2016.bib}
\end{document}